\documentclass{article}

\usepackage{soul}
\usepackage{url}
\usepackage[utf8]{inputenc}
\usepackage{graphicx}
\usepackage{amsmath}
\usepackage{amsthm}
\usepackage{booktabs} 
\usepackage{algorithm}
\usepackage{algorithmic}
\usepackage{adjustbox}
\usepackage{apxproof}
\usepackage[authoryear,square]{natbib}
\usepackage{mathptmx,stmaryrd}

\usepackage[small]{caption}
\usepackage{graphicx}
\usepackage{amsmath}
\usepackage{amsthm}
\usepackage{booktabs}
\usepackage{algorithm}
\usepackage{algorithmic}
\usepackage[switch]{lineno}
\usepackage{color, colortbl}
\usepackage{xspace}
\usepackage{enumerate,paralist}
\usepackage{subcaption}
\newenvironment{proofNestedLemma}[1][\proofname]{
  
  \begin{proof}[#1]}{\end{proof}}
  
\usepackage{amsthm}
\usepackage{tikz}
\usetikzlibrary{arrows,positioning,shapes.misc,shapes.geometric}
\usetikzlibrary{arrows.meta}   
\tikzset{>=stealth',args/.style={circle,draw=black},minimum size=10mm}

\usepackage[textsize=tiny]{todonotes}
\usepackage{amsmath}
\usepackage{amssymb}
\usepackage[normalem]{ulem}
\newtheorem{example}{Example}

\newtheorem{definition}{Definition}
\newtheorem{lemma}{Lemma}

\newtheorem{remark}{Remark}

\newtheoremrep{propositionAp}{Proposition}

\newcommand{\HD}{\mathit{HD}}
\newcommand{\IC}{\mathit{IC}}
\newcommand{\HRc}{\mathcal{HD}}
\newcommand{\HDc}{\HRc}
\newcommand{\ICc}{\mathcal{IC}}

\newcommand {\tup}[1]      {{\langle #1 \rangle}}

\newcommand{\sat}{\mathsf{sat}}
\newcommand{\dom}{\mathsf{dom}}
\newcommand{\head}{\mathsf{hd}}

\usepackage{calrsfs}
\DeclareMathAlphabet{\pazocal}{OMS}{zplm}{m}{n}
\renewcommand{\mathcal}[1]{\pazocal{#1}}

\begin{document}

\title{Operator-based semantics for choice programs: is choosing losing? (full version)}

\author{
Jesse Heyninck\\
{\small 
Open Universiteit, the Netherlands}\\
{\small University of Cape Town, South-Africa}
}

\maketitle

\begin{abstract}
Choice constructs are an important part of the language of logic programming, yet the study of their semantics has been a challenging task. So far, only two-valued semantics have been studied, and the different proposals for such semantics have not been compared in a principled way. In this paper, an operator-based framework allow for the definition and comparison of different semantics in a principled way is proposed.
\end{abstract}

\section{Introduction}
Logic programming is one of the most popular declarative formalisms, as it offers an expressive, rule-based modelling language and efficient solvers for knowledge representation. An important part of this expressiveness comes from \emph{choice constructs} \citep{simons2002extending}, that allow to state e.g.\ set constraints in the body or head of rules, and are, among others, part of the ASP-Core-2 standard \citep{calimeri2012asp}. For example, the rule $1\{p,q,r\} 2\leftarrow s.$ expresses that if $s$ is true, between 1 and 2 atoms among $p$, $q$ and $r$ are true. Choice constructs are non-deterministic, in the sense that there is more than one way to satisfy them. For example,  $1\{p,q,r\} 2$ can be satisfied by $\{p\}$, $\{p,q\}$, $\{r\}$, $\ldots$. Formulating semantics for such non-deterministic rules has proven a challenging task \citep{liu2010logic,marek2004set,faber2004recursive,son2007constructive}: several semantics have been proposed but no unifying framework for defining and comparing these semantics exists. Furthermore, attention has been restricted to two-valued semantics, in contradistinction to many other dialects for logic programming for which three- or four-valued semantics have been proposed. Moreover, many proposed semantics only allow for choice constructs in the body, but not in the head. Finally, relations with the non-deterministic construct disjunction remain unclear. 

In this paper, a unifying framework for the definition and study of semantics for logic programs with choice constructs in the head and body is provided.  This framework is based on immediate consequence operators, which are also useful for the design of explanatory tools and provide foundations for solvers \citep{kaminski2023foundations,eiter2023explaining}.
The contributions of this paper are the following: (1) we show how the famework of non-deterministic approximation fixpoint theory (AFT)  \citep{heyninck2022non}  can be used to define a wide variety of supported and stable semantics for choice logic programs, (2) we introduce the constructive stable fixpoints, allowing to (3) generalize many existing semantics for choice programs, (4) compare these semantics by introducing postulates, (5) provide a principled comparison with disjunctive logic programs.

\noindent{\bf Outline of the Paper}:
 In Section \ref{sec:back:prelim}, the background on choice programs and  non-deterministic approximation fixpoint theory is given. In Section \ref{sec:ndao:for:choice}, approximation operators for choice programs are defined. In Section \ref{sec:supported:models}, we study the resulting supported semantics. In Section \ref{sec:stable}, we define stable semantics and show representation results, while giving a postulate-based study  in Section \ref{sec:postulates}. In Section \ref{sec:disjunction}, we relate choice constructs and disjunctions. Related work and a conclusion follows in Sections \ref{sec:rel:work} and \ref{sec:conclusion}.

\noindent{\bf Relation with Previous Work}: 
 This paper extends this previous work \cite{DBLP:conf/nmr/Heyninck23a} by also considering choice construct in the body, which gives rise to different possible approximators, and comparing these operators using the notion of groundedness.

\section{Background and Preliminaries}
\label{sec:back:prelim}
We recall choice programs  and non-deterministic AFT.

\subsection{Choice Rules and Programs}\label{sec:choice:rules}
A \emph{choice atom} (relative to a set of atoms $\mathcal{A}$) is an expression $C=(\dom,\sat)$ where $\dom \subseteq \mathcal{A}$ and $\sat\subseteq \wp(\dom)$. Intuitively, $\dom$ denotes the \emph{domain} of $C$, i.e.\ the atoms relevant for the evaluation of $C$, whereas $\sat$ is the set of \emph{satisfiers} of $C$. We also denote, for $C=(\dom,\sat)$, $\dom$ by $\dom(C)$ and $\sat$ by $\sat(C)$.
For a concrete example, consider $1 \{p,q,r\} 2$ which intuitively states that between 1 and 2 of the atoms $p$, $q$ and $r$ have to be true, corresponds to the choice atom $(\{p,q,r\},\{ \{p\},\{q\},\{r\},\{p,q\},\{p,r\}, \{q,r\}\})$ (notice that $\{p,q,r\}$ is the domain and not a satisfier). For such choice atoms, we assume the domain and satisfiers are clear and can be left implicit (and similarly for constructs such as $\{p,q\}=1$ or $\{p,q\}\neq 1$).

A \emph{choice rule} has the form $C\leftarrow C_1,\ldots,C_n$ where  $C,C_1,\ldots,C_n$ are choice atoms.
 $C$ is called the \emph{head} (denoted $\head(r)$) and $C_1,\ldots,C_n$ the \emph{body}. If $C_i$ is a literal (i.e.\ $C_i=(\{\alpha\},\{\{\alpha\}\})$, abbreviated by $ \alpha$, or  $C_i=(\{\alpha\},\{\emptyset\})$, abbreviated by $\lnot \alpha$) for every $i=1,\ldots,n$, we call it a \emph{normal choice rule}. If $C$ is an atom, we call $r$ an \emph{aggregate rule}.
 A \emph{choice program} is a set of choice rules, and is called normal[aggregate] if all of the rules are so. A choice atom $C$ is \emph{monotone} if  $\dom(C)\cap x\in\sat(C)$ implies $\dom(C)\cap x'\in\sat(C)$ for any $x\subseteq x'\subseteq \mathcal{A}$, and it is \emph{convex} if for any s.t.\ $x\subseteq y\subseteq \mathcal{A}$, $\dom(C)\cap x\in\sat(C)$ and $\dom(C)\cap y\in\sat(C)$, $\dom(C)\cap z\in\sat(C)$ for any $x\subseteq z\subseteq y$.
Following \citet{liu2010logic}, a set $x\subseteq \mathcal{A}$ \emph{satisfies} a choice atom $C$ if $\dom(C)\cap x\in {\sf sat}(C)$.  An interpretation $x$ satisfies a rule $r$ if $x$ satisfies the head of $r$ or does not satisfy some choice atom in the body of $r$. $x$ is a \emph{model} of $\mathcal{P}$ if it satisfies every rule in $\mathcal{P}$. A rule $r\in\mathcal{P}$ is \emph{$x$-applicable} if $x$ satisfies the body of $r$, and the set of $x$-applicable rules in $\mathcal{P}$ is denoted by $\mathcal{P}(x)$. 
$x\subseteq \mathcal{A}$ is a  \emph{supported model} of $\mathcal{P}$ if it is a model and $x\subseteq \bigcup_{r\in \mathcal{P}(x)} \dom(\head(r))$.
For some $x\subseteq \mathcal{A}$, let 
 $\HD_\mathcal{P}(x)=\{ \head(r)\mid r\in \mathcal{P}(x)\}$ and  $\IC_\mathcal{P}(x)=$
\[\{z\subseteq \bigcup_{C\in \HD_\mathcal{P}(x)}\dom(C)\mid \forall C\in  \HD_\mathcal{P}(x):z(C)={\sf T}\}\]

\begin{example}\label{example:choice:rules:one}
Consider the program $\mathcal{P}=\{1\{p,q\} 2\leftarrow \{p,q\}\neq 1\}$. 
The choice atoms behave as follows:

\begin{center}
\begin{tabular}{l|llll}
 & $\emptyset$ & $\{p\}$ & $\{q\}$ & $\{p,q\}$\\ \hline
 $1\{p,q\}2$ & ${\sf F}$ &  ${\sf T}$ &  ${\sf T}$ &  ${\sf T}$\\
  $\{p,q\}\neq 1$ & ${\sf T}$ &  ${\sf F}$ &  ${\sf F}$ &  ${\sf T}$\\
\end{tabular}\end{center}
 This means that $\IC_\mathcal{P}(\emptyset)=\IC_\mathcal{P}(\{p,q\})=\{\{p\},\{q\},\{p,q\}\}$, whereas $\IC_\mathcal{P}(\{p\})=\IC_\mathcal{P}(\{q\})=\{\emptyset\}$. The models of $\mathcal{P}$ are $\{p\}$, $\{q\}$ and $\{p,q\}$, and only $\{p,q\}$ is supported.
\end{example}

\subsection{Approximation Fixpoint Theory}
\label{sec:AFT}
We first  recall some basic algebraic notions. %
A \emph{lattice} is a partially ordered set (poset) $\langle \mathcal{L},\leq\rangle$ s.t.\ for every $x,y\in \mathcal{L}$, a least upper bound $x\sqcup y$ and a greatest lower bound $x\sqcap y$ exist. A lattice is \emph{complete} if every $X\subseteq \mathcal{L}$ has a least upper bound $\bigsqcup X$ and a greatest lower bound $\bigsqcap X$.
$\bigsqcup \mathcal{L}$ is denoted by $\top$ and $\bigsqcap \mathcal{L}$ is denoted by $\bot$.
A function $f:X\rightarrow Y$ from a poset $\langle X,\leq_1\rangle$ to a poset $\langle Y,\leq_2\rangle$ is \emph{monotonic} if $x_1\leq_1 x_2$ implies $f(x_1)\leq_2 f(x_2)$, and $x\in X$ is a \emph{fixpoint} of $f$ if $x=f(x)$
 We define the ordinal powers of a function $f: X\rightarrow X$ as $f^0(x)=x$, $f^{\alpha+1}(x)=f(f^{\alpha}(x))$ for a successor ordinal $\alpha$, and   $f^{\alpha}(x)=\bigsqcup_{\beta<\alpha}\!f^{\beta}(x)$ for a limit ordinal $\alpha$. The following notation will be used: $[x,y]=\{z\mid x\leq z\leq y\}$. We say a pair $(x,y)$ is consistent if $x\leq y$ and total if $x=y$.
 
We now recall basic notions from non-deterministic approximation fixpoint theory (AFT) by \citet{heyninck2022non}, which generalizes approximation fixpoint theory as introduced by \citet{denecker2000approximations} to non-deterministic operators. We refer to the original paper \citep{heyninck2022non} for more details. 

 The basic idea behind non-deterministic approximation operators is that we approximate a non-deterministic operator $O$ by generating, for a given lower bound $x$ and upper bound $y$ that approximates $z$, a set of lower bounds $\{x_1,x_2,\ldots\}$ and a set of upper bounds $\{y_1,y_2,\ldots\}$ that under- respectively over-approximate $O(z)=\{z_1, z_2,\ldots\}$.
Formally, a {\em non-deterministic operator on $\mathcal{L}$} is a function $O : \mathcal{L}\rightarrow \wp(\mathcal{L}) \setminus \{\emptyset\}$. 
 For example, the operator $\IC_\mathcal{P}$ is a non-deterministic operator on the lattice $\tup{\wp(\mathcal{A}),\subseteq}$.
As the ranges of non-deterministic operators are {\em sets\/} of lattice elements, one needs a way to compare them, such as the {\em Smyth order\/} and the {\em Hoare order\/}.
Let $L = \tup{\mathcal{L},\leq}$ be a lattice, and let $X,Y \in \wp(\mathcal{L})$. Then: $X \preceq^S_L Y$ if for every $y\in Y$ there is an $x\in X$ such that $x\leq y$; and $X \preceq^H_L Y$ if for every $x\in X$ there is a $y\in Y$ such that $x\leq y$.              Given some $X_1,X_2,Y_1,Y_2\subseteq \mathcal{L}$, $X_1\times Y_1 \preceq^A_i X_2\times Y_2$ iff $X_1\preceq^S_L X_2$ and $Y_2\preceq^H_L Y_1$; and $X_1\times Y_1 \preceq^S_t X_2\times Y_2$ iff $X_1\preceq^S_L X_2$ and $Y_2\preceq^S_L Y_1$. 
The main orders, instantiated for the lattice of interest for this paper, $\langle \mathcal{A},\subseteq\rangle$ are recalled in Table \ref{tab:orders}.

Let $L=\tup{\mathcal{L},\leq}$ be a lattice.
Given an operator $\mathcal{O}:\mathcal{L}^2\rightarrow \mathcal{L}^2$, we denote by $\mathcal{O}_l$  the projection operator defined by $\mathcal{O}_l(x,y)=\mathcal{O}(x,y)_1$, and similarly for $\mathcal{O}_u(x,y)=\mathcal{O}(x,y)_2$.
 An operator $\mathcal{O}:\mathcal{L}^2\rightarrow \wp(\mathcal{L}){\setminus\emptyset}\times \wp(\mathcal{L}){\setminus\emptyset}$ 
is called a {\em non-deterministic approximating operator\/} (ndao, for short), if it is $\preceq^A_i$-monotonic (i.e.\ $(x_1,y_1)\leq_i (x_2,y_2)$ implies $\mathcal{O}(x_1,y_1)\preceq^A_i \mathcal{O}(x_2,y_2)$),
and  is \emph{exact} (i.e., for every $x\in \mathcal{L}$, $\mathcal{O}(x,x)=(\mathcal{O}_l(x,x), \mathcal{O}_{l}(x,x))$). $(x,y)$ is a fixpoint of $\mathcal{O}$ if $x\in \mathcal{O}_l(x,y)$ and $y\in \mathcal{O}_u(x,y)$.
A non-deterministic operator $O:\mathcal{L}\rightarrow \wp(\mathcal{L})$ is \emph{downward closed} if for every sequence $X=\{x_\epsilon\}_{\epsilon<\alpha}$ of elements in $\mathcal{L}$ such that:
(1) for every $\epsilon<\alpha$, $O(x_\epsilon)\preceq^S_L \{x_\epsilon\}$, and
(2) for every $\epsilon <\epsilon'<\alpha$, $x_{\epsilon'}< x_\epsilon$,
it holds that $O(\bigsqcap X)\preceq^S_L \bigsqcap X$.
We also introduce the following generalisation of an approximation operator, as it will allow us to capture some further existing semantics.
\begin{definition}
An operator $\mathcal{O}:\mathcal{L}^2\rightarrow \wp(\mathcal{L}){\setminus\emptyset}\times \wp(\mathcal{L}){\setminus\emptyset}$  is a semi-ndao iff:
(1) it is exact,
(2) $\mathcal{O}_l(\cdot,y)$ is $\preceq^S_L$-monotonic for any $y\in \mathcal{L}$, and 
(3) $\mathcal{O}_u(x,\cdot)$ is $\preceq^H_L$-anti-monotonic for any  $x\in \mathcal{L}$.
\end{definition}
Any ndao is a semi-ndao (cf.\  Lemma 1 by \citet{heyninck2022non}). A (semi-)ndao $\mathcal{O}$ \emph{approximates} an operator $O$ if $\mathcal{O}_l(x,x)=O(x,x)$ for every $x\in \mathcal{L}$. 

We recall now stable operators (given an ndao $\mathcal{O}$).
The \emph{complete lower stable operator\/} is defined by (for any $y\in \mathcal{L}$) $C(\mathcal{O}_l)(y) \ = 
 \{x \in \mathcal{L}\mid x\in \mathcal{O}_l(x,y) \mbox{ and }\lnot \exists x'< x: x'\in \mathcal{O}_l(x',y)\}$.
The \emph{complete upper stable operator\/} is defined by  (for any $x\in \mathcal{L}$) $C(\mathcal{O}_u)(x) \ =\  \{y \in \mathcal{L}\mid y\in \mathcal{O}_u(x,y) \mbox{ and }\lnot \exists y'<y:  y' \in \mathcal{O}_u(x,y')\}$.
 The \emph{stable operator\/} is given by: $S(\mathcal{O})(x,y)=(C(\mathcal{O}_l)(y),C(\mathcal{O}_u)(x) )$. $(x,y)$ is a \emph{stable fixpoint\/} of $\mathcal{O}$ if $(x,y)\in S(\mathcal{O})(x,y)$.

Other semantics (e.g.\ well-founded state, Kripke-Kleene fixpoints and state and semi-equilibrium semantics) have been defined (\citep{heyninck2022non,DBLP:journals/corr/abs-2305-10846}) and can be immediately obtained once an ndao is formulated. Due to space limitations, they are not discussed or studied here.

\section{Approximation Operators for Choice Programs}\label{sec:ndao:for:choice}
The central task is to define non-deterministic approximations of the immediate consequence operator $\IC_\mathcal{P}$. As is usual, we conceive of pairs of sets of atoms $(x,y)$ as \emph{four-valued interpretations}, where atoms in $x$ are \emph{true} whereas those in $y$ are \emph{not false}. Thus, assuming $x\subseteq y$, $(x,y)$ represents an approximation of some set $z\in[x,y]$.
The basic idea behind all the operators defined below is the same: given an input interpretation $(x,y)$, we determine a set of rules that are to be taken into account when constructing the new lower (respectively upper bound), and then take as new lower (respectively upper bounds) the interpretations that make true the heads of all these rules.  As is well-known already in the case for aggregate programs, there are various ways to  give formal substance to this idea. 
We will consider four operators inspired by previous work on aggregate or choice programs, namely $\ICc^{\sf GZ}_\mathcal{P}$ \citep{gelfond2014vicious},  $\ICc^{\sf MR}_\mathcal{P}$ \citep{marek2004set},  $\ICc^{\sf LPST}_\mathcal{P}$ \citep{liu2010logic}, and  $\ICc^\mathcal{U}_\mathcal{P}$ \citep{DBLP:journals/corr/abs-2305-10846}.
The study of further operators is left for future work.
Intuitively, the ${\sf GZ}$-operator takes into account only rules of satisfied bodies whose domain remains unchanged in $x$ and $y$. The lower bound of the ${\sf LPST}$-operator takes into account heads of rules whose body is true in every interpretation between $x$ and $y$, whereas the upper bound collects the results of applying $\IC_\mathcal{P}$ to every ``non-false'' interpretation (i.e.\ every member of $[x,y]$).
 The lower bound of the  ${\sf MR}$-operator  looks at all rules whose body is satisfied by the upper bound $y$ and by some subset of the lower bound $x$, whereas its upper bound is identical to the ${\sf LPST}$-operator upper bound. The $\mathcal{U}$-operator, finally, has as a lower and upper bound the ${\sf LPST}$-operator upper bound.
Given a choice program $\mathcal{P}$ and pair of sets of atoms $(x,y)$, let:
  \begin{align*}
  &\HDc^{{\sf GZ},l}_\mathcal{P}(x,y)=\{  C\mid \exists C\leftarrow C_1,\ldots,C_i\in \mathcal{P}, \forall i=1\ldots n:  \\
& \hspace{2.5cm} x\cap \dom(C_i)=y\cap \dom(C_i)\in\sat(C_i)  \}, \\
 &\mathcal{HD}^{{\sf LPST},l}_\mathcal{P}(x,y)=\{C\mid  \exists C\leftarrow C_1,\ldots,C_n\in \mathcal{P}, \forall i=1\ldots n: \\
&\hspace{2.5cm}\forall z\in[x,y]: z(C_i)={\sf T}\},
\\
  & \mathcal{HD}^{{\sf MR},l}_\mathcal{P}(x,y)=\{ C\mid  \exists  C\leftarrow C_1,\ldots,C_n \in \mathcal{P},  \exists z\subseteq x:\\ & \hspace{2.5cm} \forall i=1\ldots n: y(C_i)={\sf T}\mbox{ and } z(C_i)={\sf T}\},\\
\end{align*}

 We then define (for ${\sf x}\in \{{\sf LPST}, {\sf MR},{\sf GZ}\}$)  the lower bound operators by $  \ICc^{{\sf x},l}(x,y)=$
 \begin{align*}
 &\{z\subseteq \hspace*{-0.65cm} \bigcup_{ C\in \HDc^{{\sf x},l}_\mathcal{P}(x,y)}\hspace*{-0.6cm}\dom(C)\mid \forall C\in \HDc^{{\sf x},l}_\mathcal{P}(x,y):z\cap {\dom(C) \in \sat (C)}\}
\end{align*} 
Furthermore, we define (for $\dagger\in \{{\sf MR},{\sf LPST}, \mathcal{U}\}$): 
  \begin{align*}
&\ICc^{\mathcal{U},l}_\mathcal{P}(x,y) &= \ICc^{\dagger,u}_\mathcal{P}(x,y) &=\bigcup_{x\subseteq z\subseteq y}\IC_\mathcal{P}(z)
\end{align*}
 whereas $\ICc^{{\sf GZ},u}_\mathcal{P}(x,y)=\ICc^{{\sf GZ},l}_\mathcal{P}(x,y)$.
Finally, we define  (for ${\sf x}\in \{{\sf LPST}, {\sf MR},{\sf GZ},\mathcal{U}\}$):
 $$\ICc^{{\sf x}}(x,y)=(\ICc^{{\sf x},l}(x,y),\ICc^{{\sf x},u}(x,y)).$$
We also provide a handy summary of all operators in Table \ref{tab:operators}.

\begin{table*}
\begin{center}
\begin{tabular}{lll}
\rowcolor{gray!80} {\color{white}Preorder} & {\color{white}Type} & {\color{white}Definition }\\ \hline \hline
\rowcolor{gray!30}\multicolumn{3}{c}{Element Orders} \\ \hline
$\leq_i$ & $\wp(\mathcal{A})^2\times\wp(\mathcal{A})^2$ & $(x_1,y_1) \leq_i (x_2,y_2)$ iff  $x_1 \subseteq x_2$ and $y_1 \supseteq y_2$ \\
$\leq_t$ & $\wp(\mathcal{A})^2\times\wp(\mathcal{A})^2$ & $(x_1,y_1) \leq_t (x_2,y_2)$ iff $x_1 \subseteq x_2$ and $y_1 \subseteq y_2$ \\ \hline
\rowcolor{gray!30}\multicolumn{3}{c}{Set-based Orders} \\ \hline
$\preceq^S_L$ & $\wp(\wp(\mathcal{A}))\times \wp(\wp(\mathcal{A}))$ & $X \preceq^S_L Y$ iff for every $y\in Y$ there is an 
$x\in X$ s.t.\ $x\subseteq y$ \vspace{0.1cm}\\
$\preceq^H_L$ & $\wp(\wp(\mathcal{A}))\times \wp(\wp(\mathcal{A}))$ &  $X \preceq^H_L Y$ iff for every $x\in X$ there is an 
$y\in Y$ s.t.\ $x\subseteq y$ \vspace{0.1cm} \\
$\preceq^A_i$ & $\wp(\wp(\mathcal{A}))^2\times\wp(\wp(\mathcal{A}))^2$ & $(X_1,Y_1)\preceq_i^A (X_2,Y_2)$ iff $X_1\preceq^S_L X_2$ and $Y_2\preceq^H_L Y_1$ \vspace{0.1cm}\\
\end{tabular}
\end{center}
\caption{List of  the preorders used in this paper (instantiated for the lattice $\langle \mathcal{A},\subseteq\rangle$).}
\label{tab:orders}
\end{table*}

\begin{table*}
\begin{center}
{
\renewcommand{\arraystretch}{1.2}
\begin{tabular}{ll}
\rowcolor{gray!30}\multicolumn{2}{c}{Immediate Consequence Operator for choice programs} \\ \hline \hline
$\mathcal{P}(x)=$ & $\{ C\leftarrow C_1,\ldots,C_i\in \mathcal{P}\mid  \forall i=1\ldots n: x\cap \dom(C_i)\in\sat(C_i)\}$ \\
$\HD_\mathcal{P}(x)=$ & $\{ \head(r)\mid r\in \mathcal{P}(x)\}$ \\
$\IC_\mathcal{P}=$ & $\{z\subseteq \bigcup_{C\in \HD_\mathcal{P}(x)}\dom(C)\mid \forall C\in  \HD_\mathcal{P}(x):z(\head(r))={\sf T}\}$ \\ \hline
\rowcolor{gray!30} \multicolumn{2}{c}{Ndao based on \cite{gelfond2014vicious}} \\ \hline \hline
 $\HDc^{{\sf GZ},l}_\mathcal{P}(x,y)=$ & $\{  C\mid \exists C\leftarrow C_1,\ldots,C_i\in \mathcal{P}, \forall i=1\ldots n:  x\cap \dom(C_i)=y\cap \dom(C_i)\in\sat(C_i)  \}$\\
  $  \ICc^{{\sf GZ},l}(x,y)=$ & $\{z\subseteq \bigcup_{ C\in \HDc^{{\sf GZ},l}_\mathcal{P}(x,y)}\dom(C)\mid \forall C\in \HDc^{{\sf GZ},l}_\mathcal{P}(x,y):z\cap {\dom(C) \in \sat (C)}\}$\\ 
$  \ICc^{{\sf GZ},u}(x,y)=$ &  $ \ICc^{{\sf GZ},l}(x,y)$\\
$\ICc^{\sf GZ}(x,y)=$ & $( \ICc^{{\sf GZ},l}(x,y), \ICc^{{\sf GZ},u}(x,y))$\\
  \hline
    
 \rowcolor{gray!30}   \multicolumn{2}{c}{Ultimate Ndao for choice programs} \\ \hline \hline
$\ICc^{\mathcal{U},l}_\mathcal{P}(x,y)=$ & $\bigcup_{x\subseteq z\subseteq z}\IC_\mathcal{P}(z)$\\
$\ICc^\mathcal{U}(x,y)=$ & $( \ICc^{\mathcal{U},l}(x,y),  \ICc^{\mathcal{U},l}(x,y))$\\ \hline
  
\rowcolor{gray!30}  \multicolumn{2}{c}{Ndao based on \cite{liu2010logic}} \\ \hline \hline
$\HDc^{{\sf LPST},l}_\mathcal{P}(x,y)=$ & $\{C\mid \exists C\leftarrow C_1,\ldots,C_n\in \mathcal{P}, i=1\ldots n:\forall z\in[x,y]: z(C_i)={\sf T}\},$\\
  $  \ICc^{{\sf LPST},l}(x,y)=$ & $\{z\subseteq \bigcup_{ C\in \HDc^{{\sf LPST},l}_\mathcal{P}(x,y)}\dom(C)\mid \forall C\in \HDc^{{\sf LPST},l}_\mathcal{P}(x,y):z\cap {\dom(C) \in \sat (C)}\}$\\ 
$  \ICc^{{\sf LPST},u}(x,y)=$ & $\ICc^{\mathcal{U},l}(x,y)$\\
$\ICc^{\sf LPST}(x,y)=$ & $( \ICc^{{\sf LPST},l}(x,y), \ICc^{{\sf LPST},u}(x,y))$\\
  \hline
  
  \rowcolor{gray!30}  \multicolumn{2}{c}{Ndao based on \cite{marek2004set}} \\ \hline \hline
$\HDc^{{\sf MR},l}_\mathcal{P}(x,y)=$ & $\{ C\mid \exists C\leftarrow C_1,\ldots,C_n \in \mathcal{P},\exists z\subseteq x: \forall i=1\ldots n: y(C_i)={\sf T}\mbox{ and } \ z(C_i)={\sf T}\}$\\
  $  \ICc^{{\sf MR},l}(x,y)=$ & $\{z\subseteq \bigcup_{ C\in \HDc^{{\sf MR},l}_\mathcal{P}(x,y)}\dom(C)\mid \forall C\in \HDc^{{\sf MR},l}_\mathcal{P}(x,y):z\cap {\dom(C) \in \sat (C)}\}$\\ 
$  \ICc^{{\sf MR},u}(x,y)=$ & $\ICc^{\mathcal{U},l}(x,y)$\\
$\ICc^{\sf MR}(x,y)=$ & $( \ICc^{{\sf MR},l}(x,y), \ICc^{{\sf MR},u}(x,y))$\\

\end{tabular}}
\end{center}
\caption{Concrete Operators for dlps and choice programs}
\label{tab:operators}
\end{table*}

We first illustrate the behaviour of these operators:
\begin{example}
Consider again $\mathcal{P}$ from Example \ref{example:choice:rules:one}.
We will consider the three-valued interpretation $(\{p\},\{p,q\})$. 
We observe that:
\begin{itemize}
\item $\ICc^{{\sf MR},l}_\mathcal{P}(\{p\},\{p,q\})=\{ \{p\},\{q\},\{p,q\}\}$,
\item  $\ICc^{{\sf LPST},l}_\mathcal{P}(\{p\},\{p,q\})= \ICc^{{\sf GZ},\dagger}_\mathcal{P}(\{p\},\{p,q\})=\{\emptyset\}$  (for $\dagger=l,u$), and
\item $\ICc^{\mathcal{U},l}_\mathcal{P}(\{p\},\{p,q\})=\ICc^{\mathcal{U},u}_\mathcal{P}(\{p\},\{p,q\}=\{\emptyset,\{p\},\{q\},\{p,q\}\}$.
\end{itemize}
 Thus, $(\{p\},\{p,q\})$ is a fixpoint of the ${\sf MR}$- and $\mathcal{U}$-operators, but not of the ${\sf LPST}$ and ${\sf GZ}$-operators.
\end{example}

Unsurprisingly, these ndaos are not well-defined for every program (inevitably so, as this already holds for $\IC_\mathcal{P}$):
\begin{example}
Let $\mathcal{P}=\{\{p,q\}\neq 2\leftarrow; \{p,q\}=2\leftarrow\}$. For this program, no set satisfying both heads exists, and thus $\ICc^{\sf x}_\mathcal{P}(x,y)$ is not defined (for any $x,y\subseteq \mathcal{A}$). 
\end{example}
We therefore making the following

\noindent{\bf Assumption}:
We restrict attention to programs for which $\ICc_\mathcal{P}^{{\sf x},\dagger}(x,y)\neq\emptyset$ for any $x,y\subseteq \mathcal{A}$, ${\sf x}\in\{{\sf LPST}, {\sf MR},{\sf GZ},\mathcal{U}\}$ and $\dagger\in \{l,u\}$.
\vspace{0.1cm}

All operators are (semi-)approximators of $\IC_\mathcal{P}$:\footnote{Proof of all results in the paper, as well as additional results can be found in the appendix.}
\begin{propositionAprep}\label{prop:operators:are:ndaos}
  $\ICc^{\sf LPST}_\mathcal{P}$ and $\ICc^{\sf MR}_\mathcal{P}$ are ndaos for $\IC_\mathcal{P}$. $\ICc^{\sf MR}_\mathcal{P}$ is a semi-ndao for $\IC_\mathcal{P}$. $\ICc^{\sf GZ}_\mathcal{P}$ is an ndao for $\IC_\mathcal{P}$ when restricted to consistent inputs.
\end{propositionAprep}
\begin{appendixproof}
Consider an arbitrary but fixed choice program $\mathcal{P}$.
\begin{description}
\item[${\sf x}=\mathcal{U}$] Follows from Proposition 1 by \citet{DBLP:journals/corr/abs-2305-10846}.
\item[${\sf x}={\sf LPST}$]
For \emph{$\preceq^A_i$-monotonicity}, consider some $(x_1,y_1)\leq_i(x_2,y_2)$. We first show that $\dagger$: $\HDc^{l,{\sf LPST}}_\mathcal{P}(x_1,y_1)\subseteq \HDc^{l,{\sf LPST}}_\mathcal{P}(x_2,y_2)$
Suppose that $C\in \HDc^{l,{\sf LPST}}_\mathcal{P}(x_1,y_1)$, i.e.\ there is some $C\leftarrow C_1,\ldots,C_n\in \mathcal{P}$ s.t.\ for every $z\in[x_1,y_1]$ and every $i=1,\ldots,n$, $z(C_i)={\sf T}$. Then, since $[x_2,y_2]\subseteq [x_1,y_1]$, it also holds that  for every $z\in[x_2,y_2]$ and every $i=1,\ldots,n$, $z(C_i)={\sf T}$. We have thus shown that $\HDc^{l,{\sf LPST}}_\mathcal{P}(x_1,y_1)\subseteq \HDc^{l,{\sf LPST}}_\mathcal{P}(x_2,y_2)$.  
To show that   $\ICc^{{\sf LPST},l}_\mathcal{P}(x_1,y_1)\preceq^S_L\ICc^{{\sf LPST},l}_\mathcal{P}(x_2,y_2)$,
consider now some $z_2\in \ICc^{{\sf LPST},l}_\mathcal{P}(x_2,y_2) $. We show that $z_2\cap \bigcup\{\dom(C)\mid C\in\HDc^{{\sf LPST},l}_\mathcal{P}(x_1,y_1)\}\in \ICc^{{\sf LPST},l}_\mathcal{P}(x_1,y_1)$. Indeed, consider some $C\in \HDc^{{\sf LPST},l}_\mathcal{P}(x_1,y_1)$. Then $C\in \HDc^{{\sf LPST},l}_\mathcal{P}(x_2,y_2)$ (with $\dagger$) and thus $\dom(C)\cap z_2\in {\sf sat}(C)$. As $z_2\cap  \bigcup\{ \dom(C)\mid C\in\HDc^{{\sf LPST},l}_\mathcal{P}(x_1,y_1)\}\subseteq z_2$, this concludes the proof of $\preceq^S_L$-monotonicity.  
$\preceq^H_L$-monotonicity of the upper bound $\ICc^{{\sf LPST},u}_\mathcal{P}$ follows immediately from the previous item as $\ICc^{{\sf LPST},u}_\mathcal{P}=\ICc^{{\cal U},u}_\mathcal{P}$.
It can be easily observed that $\ICc^{{\sf LPST},l}_\mathcal{P}(x,x)=\ICc^{{\cal LPST},u}_\mathcal{P}(x,x)=\IC_\mathcal{P}(x)$, which demonstrates \emph{exactness} and that $\ICc_\mathcal{P}$ \emph{approximates $\IC_\mathcal{P}$}.
\item[${\sf x}={\sf MR}$] For \emph{$\preceq^S_L$-monotonicity} of $\ICc^{{\sf MR},l}_\mathcal{P}(\cdot,y)$, consider some $x_1\subseteq x_2\subseteq y\subseteq \mathcal{A}$. We show that  $\HDc^{l,{\sf MR}}_\mathcal{P}(x_1,y)\subseteq \HDc^{l,{\sf LPST}}_\mathcal{P}(x_2,y)$, the proof of $\preceq^S_L$-monotonicity is identical to that given in the previous item. Suppose thus that $C\in\HDc^{l,{\sf MR}}_\mathcal{P}(x_1,y)$, i.e.\ there is some $C\leftarrow C_1,\ldots,C_n$ s.t.\ $y(C_i)={\sf T}$ and there is some $x\subseteq x_1$ s.t.\ $x(C_i)={\sf T}$ (for every $i=1,\ldots,n$). But then also $x\subseteq x_2$ and thus $C\in \HDc^{l,{\sf MR}}_\mathcal{P}(x_2,y)$. $\preceq^H_L$-monotonicity of the upper bound was established in the previous item.  \emph{Exactness} and \emph{approximation of $\IC_\mathcal{P}$} are immediate.
\item[${\sf x}={\sf GZ}$]  For \emph{$\preceq^S_L$-monotonicity}, consider some  $(x_1,y_1)\leq_i(x_2,y_2)$ and $x_2\subseteq y_2$ (recall that we restrict the input of the ${\sf GZ}$-operator to consistent pairs). We show that $\HDc^{l,{\sf GZ}}_\mathcal{P}(x_1,y_1)\subseteq \HDc^{l,{\sf GZ}}_\mathcal{P}(x_2,y_2)$, the rest can be derived analogously to the first item. Consider thus some $C\in \HDc^{l,{\sf GZ}}_\mathcal{P}(x_1,y_1)$, which means that there is some $C\leftarrow C_1,\ldots,C_n\in\mathcal{P}$ s.t.\ $x_1\cap \dom(C_i)=y_1\cap \dom(C_i)\in \sat(C_i)$ for every $i=1,\ldots,n$. As $x_1\subseteq x_2$, we see that $x_1\cap \dom(C_i)\subseteq x_2\cap \dom(C_i)$ (and similarly we see that $y_2\cap \dom(C_i)\subseteq y_1\cap \dom(C_i)$). As we restricted ourselves to consistent pairs (and thus $x_2\subseteq y_2$), we see that $x_1\cap \dom(C_i)= x_2\cap \dom(C_i)=y_2\cap \dom(C_i)= y_1\cap \dom(C_i)$. \emph{Exactness} and \emph{approximation of $\IC_\mathcal{P}$} are immediate.
\end{description}
\end{appendixproof}

The following example shows that $\preceq^A_i$-monotonicity of $\ICc_\mathcal{P}^{\sf GZ}$ is not guaranteed when considering inconsistent interpretations:
\begin{example}
Consider $\mathcal{P}=\{p\leftarrow \{p,q\}\neq 0\}$. We see that 
\begin{eqnarray*}
&(\{p\},\{p\})\leq_i (\{p,q\},\{p\})\quad\mbox{ yet }\\
&\ICc_\mathcal{P}^{{\sf GZ},l}(\{p\},\{p\})=\{\{p\}\}\not\preceq^S_L  \ICc_\mathcal{P}^{{\sf GZ},l}(\{p,q\},\{p\})=\{\emptyset\}
\end{eqnarray*}

To see that $ \ICc_\mathcal{P}^{{\sf GZ},l}(\{p,q\},\{p\})=\{\emptyset\}$, observe that $\dom( \{p,r\}\neq 0)\cap \{p\}\neq \dom( \{p,r\}\neq 0)\cap \{p,q\}$.
\end{example}

Furthermore, the ${\sf LPST}$ and ${\sf MR}$-operators coincide for normal choice programs:
\begin{propositionAprep}\label{prop:normal:choice:MR:LPST:same}
For any normal choice program,  $\mathcal{IC}^{\sf MR}_\mathcal{P}(x,y)= \mathcal{IC}^{\sf LPST}_\mathcal{P}(x,y)$. 
\end{propositionAprep}
\begin{appendixproof}
We have to show that $\mathcal{HD}^{l,{\sf MR}}(x,y)=\mathcal{HD}^{l,{\sf LPST}}(x,y)$ for a normal choice program $\mathcal{P}$ (the upper bound is already identical by definition), which can be seen by observing that for any atom $\alpha$, $z(\alpha)={\sf T}$ for every $z\in[x,y]$ iff $ \alpha\in x$, and for any negated literal $\lnot \alpha$, $z(\lnot \alpha)={\sf T}$ for every $z\in[x,y]$ iff $\alpha\not\in y$.
\end{appendixproof}

\begin{remark}\label{remark:semanticsNLPs:are:preserved}
It can be easily verified that the  ${\sf LPST}$ and ${\sf MR}$-operators coincide with the four-valued Przymusinski operator for normal logic programs \citep{przymusinski1990well}, which implies, with Proposition 3, Proposition 11 and Theorem 6 by \citet{heyninck2022non}, that the semantics based on these operators faithfully generalize the (partial) supported, (partial) stable and well-founded semantics of normal logic programs.
\end{remark}

\section{Supported Model Semantics}\label{sec:supported:models}
In this section, we look at the fixpoints of $\ICc^{\sf x}_\mathcal{P}$, which give a three-valued generalisation of the supported model semantics by \citet{liu2010logic} (and the three-valued model semantics of normal logic programs). Intuitively, fixpoints of $\ICc^{\sf x}_\mathcal{P}$ generalise the idea that only atoms supported by activated rules can be accepted.

A first insight is that the total fixpoints of all operators coincide with \emph{supported models}  \citep{liu2010logic}.
\begin{propositionAprep}\label{prop:supported}
Let a choice program $\mathcal{P}$ and ${\sf x}\in\{{\sf LPST}, {\sf MR},{\sf GZ},\mathcal{U}\}$  be given. Then $(x,x)$ is a fixpoint of $\ICc^{{\sf x}}_\mathcal{P}$ iff $x$ is a supported model of $\mathcal{P}$.
\end{propositionAprep}
\begin{appendixproof}
Notice first that, as $ \ICc^{{\sf x}}_\mathcal{P}$ approximates $\IC_\mathcal{P}$ (Proposition \ref{prop:operators:are:ndaos}),  $\ICc^{{\sf x}}_\mathcal{P}(x,x)=\IC_\mathcal{P}(x)$ for any $x\subseteq \mathcal{A}$.
For the $\Rightarrow$-direction, suppose that $x\in\ICc^{{\sf x},l}_\mathcal{P}(x,x)= \IC^{c}_\mathcal{P}(x)$. We first show $x$ is a model of $\mathcal{P}$. Indeed, consider a rule $ C\leftarrow C_1,\ldots,C_n$ s.t.\ $x$ satisfies $C_i$ for $i=1,\ldots,n$. Then $x(C_i)={\sf T}$ and thus $C\in \HDc^{{\sf x},l}_\mathcal{P}(x,x)=\HD_\mathcal{P}(x)$, which implies, since $x\in  \IC^{c,l}_\mathcal{P}(x,x)$, that  $x\cap \dom(C)\in \sat(C)$. That $x$ is supported follows immediately from the fact that  $x\in \IC_\mathcal{P}(x)$ implies $z\subseteq \bigcup_{ C\in \HD^{l}_\mathcal{P}(x)}\dom(C)$.
The $\Leftarrow$-direction is analogous.
\end{appendixproof}
However, our extension to three-valued semantics allows to give semantics to a wider class of programs:
\begin{example}
Let $\mathcal{P}=\{ \{p,q\}=1\leftarrow \{p\}\neq 1; p\leftarrow q. \}$. It can be easily checked that this program has no two-valued supported model. However, $(\emptyset,\{p\})$ is a fixpoint of $\ICc^{\sf x}_\mathcal{P}$ for $x\in \{{\sf LPST},{\sf MR},\mathcal{U}\}$.
\end{example}
A more detailed investigation of conditions for the existence of fixpoints  is left for future work.

We now show that fixpoints of all four operators satisfy a notion of \textit{supportedness}, relative to how rule bodies are evaluated according to their definition. For example, $C$ can be said to be true in $(x,y)$ according to the ${\sf LPST}$-operator if $z(C)={\sf T}$ for every $z\in [x,y]$. We formalize this as follows. Given $(x,y)$ and $\mathcal{IC}^{\sf x}_\mathcal{P}$, $a\in x$ supported if $a$ occurs in the domain of the head of a rule whose body is true, i.e.\ there is some $C\leftarrow C_1,\ldots,C_n$ s.t.\ $a\in\dom(C)$, and if would replace $C$ by the dummy atom $p$, $p$ is the only consequence of $\ICc^{{\sf x},l}_{\{p\leftarrow C_1,\ldots,C_n\}}(x,y)$ (and similarly for $y$).
\begin{propositionAprep}\label{proposition:fixpoints:are:supported}
Let a choice program $\mathcal{P}$ and ${\sf x}\in\{{\sf LPST}, {\sf MR},{\sf GZ},\mathcal{U}\}$  be given. Then $(x,y)\in \ICc^{\sf x}_\mathcal{P}(x,y)$ implies that for every $a\in y$, there is some $C\leftarrow C_1,\ldots,C_n$ with $a\in\dom(C)$ s.t.\ 
 $\ICc^{{\sf x}}_{\{p\leftarrow C_1,\ldots,C_n\}}(x,y)=(\{\{p\}\cap x\},\{\{p\}\cap y\})$.
\end{propositionAprep}
\begin{appendixproof}
We show this for ${\sf x}=\mathcal{U}, {\sf LPST}$, the proof of the other cases are analogous.
\begin{description}
\item[${\sf x}=\mathcal{U}$] Suppose that $(x,y)\in \ICc^\mathcal{U}_\mathcal{P}(x,y)$ and $p\in y$ (and thus, also $p\in x$ as $x=y$ by definition of $\ICc^\mathcal{U}_\mathcal{P}$).  This means that there is some $z\in [x,y]$ s.t.\ $p\in \dom(C)$ for some $C\leftarrow \phi$ for which $z(\phi)={\sf T}$. But then $\ICc^\mathcal{U}_{\{p\leftarrow \phi\}}(x,y)=(\{\{p\}\},\{\{p\}\})$.

\item[${\sf x}={\sf LPST}$] Suppose that $(x,y)\in \ICc^{\sf LPST}_\mathcal{P}(x,y)$ and $p\in y$. The case for the upper bound follows from the previous case. Suppose thus that $p\in x$.
By definition of $\ICc^{\sf LPST}$,
$p\in \bigcup_{C\in \HDc_\mathcal{P}^{{\sf LPST},l}(x,y)}\dom(C)$, which means that there is some $C\leftarrow C_1,\ldots,C_n$ s.t.\ for every $i=1,\ldots,n$, and for every 
for every $z\in[x,y]$, $z(C_i)={\sf T}$. But then  $\ICc^{{\sf x},l}_{\{p\leftarrow C_1,\ldots,C_n\}}(x,y)=\{\{p\}\}$.
\end{description}
\end{appendixproof}

\paragraph*{Normal choice programs}
We can simplify the definition of supported models for normal choice programs. In order to do this, we first have to generalize the four-valued truth-assignments to choice constructs. The following forms a generalization of assignment of truth-values to choice constructs that forms a natural generalization of the assignment of atoms to choice constructs.
We first recall the bilattice ${\sf FOUR}$,  
consisting of the elements ${\sf T}$ (true), ${\sf F}$ (false), ${\sf U}$ (undecided) and ${\sf C}$ (contradictory) and two order relations $\leq_i$ and $\leq_t$:
{\begin{center}
\resizebox{0.2\textwidth}{!}
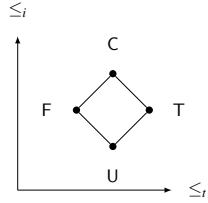
{%
\begin{tikzpicture}[node distance=1cm, auto, >=latex, scale=0.5]
		\node (a) {};
		\node (b) [above of=a, yshift=2.5cm] {$\leq_i$};
		\node (c) [right of=a, xshift=2.5cm] {$\leq_t$};
		\draw[->] (a.center) -- (b);
		\draw[->] (a.center) -- (c);
		\tikzstyle{dot}= [circle, fill, minimum size=4pt,inner sep=0pt, outer sep=0pt]
		\node (third) [dot,above right of=a, node distance=1.2cm, xshift=1cm, label=below:{${\sf U}$}] {};
		\node (second) [dot,above left of=third, label=left:{{${\sf F}$}}] {};
		\node (fourth) [dot,above right of=third, label=right:{${\sf T}$}] {};
		\node (first) [dot, above right of= second,label=above:{{${\sf C}$}}] {};	

		\path[color=black] (second) edge (first);
		\path[color=black] (third) edge (second);
		\path[color=black] (fourth) edge (third);
		\path[color=black] (first) edge (fourth);
	\end{tikzpicture}}
	\end{center}}
\begin{definition}
Given a choice construct $C$ and an interpretation $(x,y)$, we say that:
 \begin{itemize}
\item ${(x,y)(C)={\sf T}}$ if ${x\cap \dom(c)\in \sat (C)}$ and $y\cap \dom(c)\in \sat (C)$;
\item ${(x,y)(C)={\sf F}}$ if ${x\cap \dom(c)\not\in \sat (C)}$ and $y\cap \dom(c)\not\in \sat (C)$; 
\item $(x,y)(C)={\sf C}$ if $x\cap \dom(c)\in \sat (C)$ and $y\cap \dom(c)\not\in \sat (C)$,
\item $(x,y)(C)={\sf U}$ if $x\cap \dom(c)\not\in \sat (C)$ and $y\cap \dom(c)\in \sat (C)$.
\end{itemize}
\end{definition}
We define \emph{three-valued models} of a normal choice program $\mathcal{P}$ as consistent interpretations $(x,y)$ for which $\bigsqcap_{\leq_t}\{(x,y)(C_i)\mid i=1\ldots n\}\leq_t (x,y)(C)$ for every $C\leftarrow C_1,\ldots,C_n \in \mathcal{P}$, and \emph{supported}\footnote{These models have been called \emph{weakly supported models} \citep{brass1995characterizations} for disjunctive logic programs.} models as models $(x,y)$ of $\mathcal{P}$ s.t.\ for every $p \in  y$, there is a rule $C\leftarrow C_1,\ldots,C_n\in \mathcal{P}$ with $p\in\dom(C)$ and $\bigsqcap_{\leq_t}\{(x,y)(C_i)\mid i=1\ldots n\}\geq_t (x,y)(p)$. I.e., a model is supported if for every non-false atom $p$, we have a reason in the form of an activated rule  for accepting (or not rejecting) that atom.

Three-valued supported models of $\mathcal{P}$ coincide with fixpoints of $\ICc^{\sf LPST}_\mathcal{P}$ and $\ICc^{\sf MR}_\mathcal{P}$, whereas those of $\ICc^{\sf GZ}_\mathcal{P}$ are a subset of the three-valued supported models:
\begin{propositionAprep}
Let some normal choice program $\mathcal{P}$ and ${\sf x}\in\{{\sf LPST},{\sf MR}\}$ be given.  Then  $(x,y)$ is a three-valued supported model of $\mathcal{P}$ iff $(x,y)$ is a fixpoint of $\ICc^{\sf x}_\mathcal{P}$. Furthermore, if $(x,y)$ is a fixpoint of $\ICc^{\sf GZ}_\mathcal{P}$ then $(x,y)$ is supported, but not always vice-versa. Supported models might not be fixpoints of $\ICc_\mathcal{P}^\mathcal{U}$ and vice-versa.
\end{propositionAprep}
\begin{appendixproof}
Let $\mathcal{P}$ be an arbitrary but fixed normal choice program.

\begin{description}
\item[${\sf x}={\sf LPST},{\sf MR}$:]
For the $\Rightarrow$-direction, suppose that $(x,y)$ is a three-valued supported model of $\mathcal{P}$. As for every $p \in  y$,   there is a rule $C\leftarrow \phi\in \mathcal{P}$ such that $p\in\dom(C)\cap x$ and $(x,y)(\phi)\geq_t (x,y)(p)$, we see that $x\subseteq \bigcup_{C\in \HDc^{{\sf x},l}_\mathcal{P}(x,y)}\dom(C)$ (and similarly for $y$). That for every $C\leftarrow \phi \in\HDc^{c,l}_\mathcal{P}(x,y)$, $\dom(C)\cap x\in \sat(C)$ follows from $(x,y)$ being a model of $\mathcal{P}$ (and similarly for $y$).
For the $\Leftarrow$-direction, suppose $(x,y)\in\ICc^{\sf x}_\mathcal{P}(x,y)$. That $(x,y)$ is a model follows straightforwardly from the definition of $\ICc^{\sf x}_\mathcal{P}(x,y)$. Consider now some $p\in y$. As $y\subseteq \bigcup_{C\in \HDc^{{\sf x},u}_\mathcal{P}(x,y)}\dom(C)$, there is some $C\leftarrow \bigwedge_{i=1}^n\alpha_i\land \bigwedge_{i=1}^m\lnot \beta_i\in \mathcal{P}$ s.t.\ $p\in \dom(C)\cap y$ (and similarly for $x$) and $\alpha_i\in y$ for $i=1\ldots n$ and $\beta_j\not\in x$ for $j=1\ldots m$.

\item[${\sf x}={\sf GZ}$:] Suppose that $(x,y)\in\ICc^{\sf GZ}_\mathcal{P}(x,y)$. That $(x,y)$ is a model follows straightforwardly from the definition of $\ICc^{\sf GZ}_\mathcal{P}(x,y)$.  Consider now some $p\in y$. As $y\subseteq \bigcup_{C\in \HDc^{{\sf GZ},u}_\mathcal{P}(x,y)}\dom(C)$, there is some $C\leftarrow \bigwedge_{i=1}^n\alpha_i\land \bigwedge_{i=1}^m\lnot\beta_i\in \mathcal{P}$ s.t.\ $p\in \dom(C)\cap y$ (and similarly for $x$) and $\alpha_i\in  y$ for $i=1\ldots n$ and $\beta_j\not\in x$ for $j=1\ldots m$. Thus $(x,y)$ is a supported model.

Supported models might not be fixpoints of the ${\sf GZ}$-operator:
For this, consider $\mathcal{P}=\{p\leftarrow \lnot q; q\leftarrow \lnot p\}$. Then $(\emptyset,\{p,q\})$ is a supported model. Yet as $\emptyset\cap \{q\}\neq \{p,q\}\cap \{q\}$ (and similarly for $p$), $\ICc^{\sf GZ}_\mathcal{P}(\emptyset,\{p,q\})=(\emptyset,\emptyset)$.

\item[${\sf x}=\mathcal{U}$:]   Supported models might not be fixpoints of the $\mathcal{U}$-operator:
Consider the program $\mathcal{P}=\{p\leftarrow q; q\leftarrow \lnot q\}$. Then $(\emptyset,\{p,q\})$ is a supported model. However, as $\bigcup_{z\subseteq \{p,q\}}\IC_\mathcal{P}(z)=\{ \{p,q\}\}$, we see that $\emptyset\not\in \ICc^{l,\mathcal{U}}_\mathcal{P}(\emptyset,\{p,q\})$. 

Fixpoints of the $\mathcal{U}$-operator might not be supported models:
Consider the program $\mathcal{P}=\{ 1\{p,q\}\leftarrow p,\lnot q\}$. Then $\ICc_\mathcal{P}^\mathcal{U}(\{p\},\{p,q\})=(\{\{p\},\{q\},\{p,q\}\},\{\{p\},\{q\},\{p,q\}\})$ and thus $(\{p\},\{p,q\})$ is a fixpoint. However, it is not supported, as $(\{p\},\{p,q\})(\lnot q)={\sf U}$ yet $(\{p\},\{p,q\})(p)={\sf T}$ (i.e.\ we cannot find any support for $p$ being true).
\end{description}
\end{appendixproof}

\paragraph*{Models as pre-fixpoints}
It is well-known that for many non-monotonic formalisms, pre-fixpoints of an operator can characterise models of the corresponding knowledge base. For the general case of choice constructs, this correspondence does not hold:
\begin{example}
Consider $\mathcal{P}=\{{\{p,q\}=1\leftarrow}\}$. Then $(\{p,q\},\{p,q\})$ is a pre-fixpoint of $\ICc_\mathcal{P}$ (as $\ICc_\mathcal{P}(\{p,q\},\{p,q\})=\{\{p\},\{q\}\}\times \{\{p\},\{q\}\}\preceq^S_t (\{p,q\},\{p,q\})$) yet it is not a model (as the only models are $\{p\}$ and $\{q\}$): there is no way to prove $p$ and $q$.
\end{example}
For choice programs with monotone heads, this correspondence \emph{does} hold:
\begin{propositionAprep}
Let some choice program $\mathcal{P}$ s.t.\ for every $C\leftarrow C_1,\ldots,C_n$, $C$ is monotone and some ${\sf x}\in \{ {\sf LPST}, {\sf MR}, {\sf GZ}, \mathcal{U}\}$ be given. Then  $x$ is a model of $\mathcal{P}$ iff $\ICc^{l,{\sf x}}_\mathcal{P}(x,x)\preceq^S_L x$. 
\end{propositionAprep}
\begin{appendixproof}
Suppose that $\mathcal{P}$ is a choice program with monotone choice constructs in the head. For the $\Rightarrow$-direction, suppose that $x$ is a model of $\mathcal{P}$. 
We show that $x\cap \bigcup_{C\in\HD(x)_\mathcal{P}(x)}\dom(C)\in \IC_\mathcal{P}(x)=\ICc^{{\sf x},l}_\mathcal{P}(x,x)$ for any ${\sf x}\in \{ {\sf LPST}, {\sf MR}, {\sf GZ}, \mathcal{U}\}$ (Notice that $ \IC_\mathcal{P}(x)=\ICc^{{\sf x},l}_\mathcal{P}(x,x)$ holds in view of Proposition \ref{prop:operators:are:ndaos}), which implies $\ICc^{{\sf x},l}_\mathcal{P}(x,x)\preceq^S_t x$. Indeed, consider some $C\in\HD(x)_\mathcal{P}(x)$, i.e.\ there is some $C\leftarrow C_1,\ldots,C_n \in \mathcal{P}$ with $\dom(C_i)\cap x\in \sat(C_i)$ for $i=1,\ldots,n$. As $x$ is a model, also $\dom(C)\cap x\in \sat(C)$. Thus $x\cap \dom(C)\in \IC_\mathcal{P}(x)$. The $\Leftarrow$-direction is similar.
\end{appendixproof}

\section{Stable Semantics}\label{sec:stable}
We now move to the stable semantics, whose main aim is favvoding the acceptance of self-supporting cycles, e.g.\ accepting $(\{p\},\{p\})$ as a reasonable model of the program $\{p\leftarrow p\}$. For the deterministic case, this is done by looking at fixpoints of the \emph{stable operator}, obtained by calculating a new lower bound as the least fixpoint of $\mathcal{O}_l(\cdot,y)$, where $y$ is the input upper bound (and similarly for the upper bound). Intuitively, we take the least information the upper bound obliges use to derive.  In contradistinction to the deterministic case, there are divergent options for how to define stable semantics for non-deterministic operators.
We first consider the minimality-based stable semantics known from non-deterministic AFT \citep{heyninck2022non}, defined as the $\leq$-minimal fixpoints of $\mathcal{O}_l(.,y)$ (cf.\ Section \ref{sec:AFT}), which where shown to generalize the (partial) stable model semantics for disjunctive (aggregate) programs  \citep{heyninck2022non,DBLP:journals/corr/abs-2305-10846}. For choice constructs, this construction is overly strong:
\begin{example}\label{example:choice:rules:stable}
Consider the program $\mathcal{P}=\{ 1 \{p,q\} 2\leftarrow \}$. Intuitively, this rule allows to choose between one and two among $p$ and $q$. The stable  version of $\ICc^{\sf x}_\mathcal{P}$ (cf.\ Section \ref{sec:choice:rules}) behaves as follows (for any ${\sf x}={\sf LPST},{\sf MR},{\sf GZ}$):
\[S(\ICc^{{\sf x},l}_\mathcal{P})(x)=\{\{p\},\{q\}\} \mbox{ for any }x\subseteq \mathcal{A}\]
 $\{p\}$ and $\{q\}$ are the two stable fixpoints of $\ICc^{\sf x}_\mathcal{P}$. This is undesirable, as according to the intuitive reading of $\mathcal{P}$, $\{p,q\}$ should also be allowed as a stable interpretation.
\end{example}
If we take one step back, we can explain the choice for minimal fixpoints, and their shortcomings in the context of choice constructs, in stable non-deterministic operators \citep{heyninck2022non} as follows. For deterministic operators, the stable version of an approximation operator $\mathcal{O}$ is defined as the greatest lower bound (glb) of fixpoints of $\mathcal{O}_l(.,y)$. For deterministic operators over finite lattices, the minimal fixpoint of $\mathcal{O}_l(.,y)$ are identical to the  glb of fixpoints of $\mathcal{O}_l(.,y)$, and it is also identical to the fixpoint obtained by iterating $\mathcal{O}_l(.,y)$ starting from $\bot$ (i.e.\ $\bigcup_{i=1}^\infty \mathcal{O}_l^i(\bot,y)$). For non-deterministic operators, this correspondence does not hold. Indeed, the glb of fixpoints of $\mathcal{O}_l(.,x)$ is often too weak (e.g.\  for the program $\mathcal{P}$ from Example \ref{example:choice:rules:stable} we get $\{p\}\cap \{q\}\cap\{p,q\}=\emptyset$ as the glb of fixpoints). However, this still leaves a third choice: namely looking at fixpoints reachable by applications of $\mathcal{O}_l(.,y)$ starting from $\bot$. 
E.g.\ for choice programs, we are interested in the fixpoints of $\ICc^{{\sf x},l}_\mathcal{P}(.,y)$ that  can built them up from the ground up (i.e.\ from $\emptyset$) by a sequence of applications of $\ICc^{{\sf x},l}_\mathcal{P}(.,y)$. 
We first generalize the notion of a well-founded sequence by \citet{denecker2014well}:
\begin{definition}\label{def:well-founded:sequence}
Given a non-deterministic operator $O:\mathcal{L}\to \wp(\mathcal{L})$ over a complete lattice, a sequence $x_0,\ldots,x_n\subseteq \mathcal{L}$ is \emph{well-founded relative to $O$} if: (1) $x_0=\bot$;
(2) $x_i\leq x_{i+1}$ and $x_{i+1}\in O(x_i)$ for every successor ordinal $i\geq 0$; and (3)  $x_\lambda=\mathrm(\bigsqcap\{x_i\}_{i<\lambda})$ for a limit ordinal $\lambda$.
The well-founded sequences relative to $O$ are denoted by ${\sf wfs}(O)$.
\end{definition}

\begin{remark}
The assumption of a complete lattice in  Definition \ref{def:well-founded:sequence} is needed since the greatest lower bound is used in point 3 of Definition \ref{def:well-founded:sequence}.
\end{remark}

Notice that, in contradistinction to the deterministic version of a well-founded sequence \citet{denecker2014well}, we require not merely that  $x_{i+1}\preceq^S_L O(x_i)$ (or, in the case of deterministic operators $O$, $x_{i+1}< O(x_i)$) but  $x_{i+1}\in O(x_i)$. This is to ensure that $x_{i+1}$ can actually be constructed from $x_i$. For non-deterministic operators, this is not ensured by $x_{i+1}\preceq^S_L O(x_i)$:
\begin{example}
Let $\mathcal{P}=\{\{p,q\} = 2\leftarrow \}$. If we would allow for $x_{i+1}\preceq^S_L O(x_i)$ in Definition \ref{def:well-founded:sequence}.(2), $\emptyset, \{p\}$ would be a well-founded sequence according to $\ICc^{l}_\mathcal{P}(\cdot,y)$ (for any $y\subseteq \{p,q\}$) as $\{p\}\preceq^S \ICc^{l}_\mathcal{P}(\{p\},y)=\{\{p,q\}\}$. However, we have no way of deriving just $p$ from the program $\mathcal{P}$.
\end{example}

We now define the \emph{constructive stable operator}:
\begin{definition}
Given an semi-ndao $\mathcal{O}$ over a complete lattice $\mathcal{L}$ with $y\in\mathcal{L}$, the \emph{c(onstructive)-complete lower bound operator} is defined as:
\[C^c(\mathcal{O}_l)(y)=\{x\in \mathcal{O}_l(x,y)\mid \exists x_0,..,x\in {\sf wfs}(\mathcal{O}_l(.,y))\}\]
The \emph{c-complete upper bound operator} $C^c(\mathcal{O}_u)$ is defined analogously, and the \emph{c-stable operator} is defined as $S^c(\mathcal{O})(x,y)=(C^c(\mathcal{O}_l)(y), C^c(\mathcal{O}_u)(x))$.
 $(x,y)$ is a \emph{c-stable fixpoint} iff $(x,y)\in S^c(\mathcal{O})(x,y)$.
\end{definition}

\begin{example}[Example \ref{example:choice:rules:stable} continued]\label{example:choice:rules:stable:2}
Consider again the program from Example \ref{example:choice:rules:stable}. We see that (for any $y\subseteq \{p,q\}$), $S^c(\ICc^{{\sf x},l}_\mathcal{P})(y)=\{\{p\},\{q\},\{p,q\}\}$, as $\emptyset,\{p\}$, $\emptyset,\{q\}$ and $\emptyset,\{p,q\}$ are all well-founded sequences. Thus, the total c-stable fixpoints of $\IC^{{\sf x}}_\mathcal{P}$ are $(\{p\},\{p\})$, $(\{q\},\{q\})$ and $(\{p,q\},\{p,q\})$ (for ${\sf x}={\sf LPST},{\sf MR},{\sf GZ},\mathcal{U}$).
\end{example}

It is easy to see that for a deterministic operator over a complete lattice, the c-stable operator coincides with the stable operator known from deterministic AFT.
The c-stable operator  generalizes the minimality-based one:
\begin{propositionAprep}\label{prop:stable:operator:is:subset:of:constructive:stable:operator}
Let an (semi-)ndao $\mathcal{O}$ over a complete lattice $L=\langle\mathcal{L},\leq\rangle$ be given s.t.\ $\mathcal{O}_u(x,\cdot)$ is $\preceq^S_L$-monotonic for any $x\in \mathcal{L}$. Then $C^c(\mathcal{O}_\dagger)(y)\supseteq C(\mathcal{O}_\dagger)(y)$ for any $y\in \mathcal{L}$ and $\dagger=l,u$. Furthermore,  if  $(x,y)\in S(\mathcal{O})(x,y)$ then  $(x,y)\in S^c(\mathcal{O})(x,y)$.
\end{propositionAprep}
\begin{appendixproof}
We first show that $(\dagger)$: for any $\preceq^S_L$-monotonic operator $O$, for every $x\in \min_\leq \{x\in \mathcal{L}\mid x\in O(x)\}$, there is a well-founded sequence $x_0,\ldots,x_n,x$ relative to $O$.
Indeed, as $O$ is $\preceq^S_L$-monotonic, for any $x_1\leq x$, $O(x_1)\preceq^S_L O(x)\ni x$, i.e.\ there is some $x_2\in O(x_1)$ with $x_2\leq x$. This means we can define a $\leq$-monotonic operator $O_s: \{x\in \mathcal{L}\mid x'\leq x\}\rightarrow \{x'\in \mathcal{L}\mid x'\leq x\}$ by simply defining a selection function that selects some $x_2\in O(x_1)$ s.t.\ $x_2\leq x$ for every $x_1\leq x$. We now show that $\{x '\in \mathcal{L}\mid x'\leq x\}$ is a complete partially ordered set. Indeed, consider some $X\subseteq \{x'\in \mathcal{L}\mid x'\leq x\}$. As $L$ is a complete lattice, $\bigsqcup X\in \mathcal{L}$. As every $x'\in X$, $x'\leq x$, also $\bigsqcup X\leq x$, thus $\bigsqcup X\in \{x'\in \mathcal{L}\mid x'\leq x\}$. 
As $\{x '\in \mathcal{L}\mid x'\leq x\}$ is a complete partially ordered set and $O_s$ is a monotone mapping, with 
Theorem 5.1 of \citet{cousot1979constructive}, we know that there is a unique well-founded sequence $O_s^1(\bot),O_s^2(\bot)\ldots x'$ that results in a least fixpoint $x'$ of $O_s$.
 Clearly also $x'$ is a fixpoint of $O$. As $x$ is a minimal fixpoint of $O$ and $x'\leq x$, $x=x'$.

We can now show the first part of the proposition. 
We first show that $C^c(\mathcal{O}_l)(y)\supseteq C(\mathcal{O}_l)(y)$: since $\mathcal{O}$ is $\preceq^A_i$-monotonic, with Lemma 2 by \citet{heyninck2022non} respectively the definition of a semi-ndao, $\mathcal{O}_l(.,y)$ is $\preceq^S_L$-monotonic, which with $\dagger$ implies $C^c(\mathcal{O}_l)(y)\supseteq C(\mathcal{O}_l)(y)$. That $C^c(\mathcal{O}_u)(y)\supseteq C(\mathcal{O}_u)(y)$ follows with the assumption that $\mathcal{O}_u(x,\cdot)$ is $\preceq^S_L$-monotonic and $\dagger$. %

For the second part, observe that if $(x,y)\in S(\mathcal{O})(x,y)$ then with the first part, $x\in  S^c(\mathcal{O}_l)(y)$ and $y\in  S^c(\mathcal{O}_u)(x)$ and thus $(x,y)\in S^c(\mathcal{O})(x,y)$.
\end{appendixproof}

Thus, what might appear to be a change to the previously formulated stable semantics for non-deterministic approximation operators \citep{heyninck2022non}, is a mere generalization of these semantics.

In general, non-deterministic operators might not admit a fixpoint, which means that the c-complete operator is not always well-defined (e.g.\ $C^c(O_l)(x)=\emptyset$ or $C^c(O_u)(y)=\emptyset$). We illustrate this for $C^c(O_u)(y)$, by giving an example of a $\preceq^H_L$-monotonic operator that does not admit a well-founded sequence.
\begin{example}\label{example:counter:upwards:closed}
 Let $O:\mathbb{N}\cup \{\infty\}$ where $O(i)=\{i+1\}$ for any $i\in \mathbb{N}$ and $\mathbb{\infty}=\mathbb{N}$. Observe that $O$ is $\preceq^H_L$-monotonic, as for any finite $i,j$ s.t.\ $i\leq j$, $O(i)=\{i+1\}$ and $i+1\leq j+1 \in O(j)$, and for any finite $i$, $i\in O(\infty)$. However, it is clear that $O$ admits no fixpoint and thus no well-founded sequence.
\end{example}
To ensure well-definedness of the c-stable operator
we will assume that the lower bound is downwards closed and the upper bound satisfies the following, analogous, notion, called \emph{upwards closedness}:
\begin{definition}
An operator $O$ is \emph{upwards closed} if for every sequence $X=\{x_\epsilon\}_{\epsilon<\alpha}$ of elements in $\mathcal{L}$ s.t.\
\begin{enumerate}
\item for every $\epsilon<\alpha$, $x_\epsilon\preceq^H_L O(x_\epsilon)$, and
\item for every $\epsilon' <\epsilon<\alpha$, $x_{\epsilon'}<x_\epsilon$,
\end{enumerate}  
it holds that $\bigsqcup X\preceq^H_L O(\bigsqcup X)$.
\end{definition}

It can be easily observed that the operator from Example \ref{example:counter:upwards:closed} is not upwards closed. Downwards closedness of $\mathcal{O}_l(\cdot,y)$ and upwards closedness of $\mathcal{O}_l(x,\cdot)$ guarantee well-definedness of the c-stable operator:\nocite{kuratowski1922methode,zorn1935remark}
\begin{propositionAprep}
For any (semi-)ndao $\mathcal{O}$ over a complete lattice s.t.\ $\mathcal{O}_l(.,y)$ is downwards closed and $\mathcal{O}_u(x,.)$ is upwards closed for any $x,y\in \mathcal{L}$, $C^c(\mathcal{O}_l)(y)\neq\emptyset$ and $C^c(\mathcal{O}_y)(x)\neq\emptyset$.
\end{propositionAprep}
\begin{appendixproof}
We first show that $C^c(\mathcal{O}_l)(y)\neq\emptyset$.
Suppose first that $\mathcal{O}$ is an ndao. Then the proposition is immediate as $C(\mathcal{O}_l)(y)\neq\emptyset$ and $C(\mathcal{O}_y)(x)\neq\emptyset$ in view of Proposition 13 by \citet{heyninck2022non} and $C^c(\mathcal{O}_l)(y)\supseteq C(\mathcal{O}_l)(y)$ and $C^c(\mathcal{O}_u)(x)\supseteq C(\mathcal{O}_u)(x)$ (Proposition \ref{prop:stable:operator:is:subset:of:constructive:stable:operator}). 
In fact, it can be checked that the proof of Proposition 13 by \citet{heyninck2022non}, which means the proofs generalize to semi-ndaos.

We now show that $C^c(\mathcal{O}_y)(x)\neq\emptyset$. For this, we show some intermediate results (Lemma \ref{lemma:maximal-post-fixpoint:also:fixpoint} and Lemma \ref{prop:h:monotonic:has:a:fp}). It will be useful to define, given a non-deterministic operator $O$ over $\mathcal{L}$, a post-fixpoint of $O$ is any $x\in\mathcal{L}$ s.t.\ $x\preceq^H_L O(x)$. 
\begin{lemma}\label{lemma:maximal-post-fixpoint:also:fixpoint}
Let $O$ be an upwards closed $\preceq^H_L$-monotonic operator over a complete lattice. Then any $\leq$-maximal post-fixpoint of $O$ is also a fixoint of $O$.
\end{lemma}
\begin{proofNestedLemma}
Suppose that $x$ is a $\leq$-maximal post-fixpoint of $O$. Thus, $\{x\}\preceq^H_L O(x)$, i.e.\ there is some $z\in O(x)$ s.t.\ $x\leq z$. As $O$ is $\preceq^H_L$-monotonic, this implies $O(x)\preceq^H_L O(z)$, and, as $z\in O(x)$, this means that $\{z\}\preceq^H_L O(z)$, which means that $z$ is a post-fixpoint. As $x$ is a $\leq$-maximal post-fixpoint and $x\leq z$, this implies $z=x$. Thus, $x\in O(x)$ which means $x$ is a fixpoint of $O$.
\end{proofNestedLemma}

\begin{lemma}\label{prop:h:monotonic:has:a:fp}
Any upwards closed $\preceq^H_L$-monotonic operator over a complete lattice admits a fixpoint. 
\end{lemma}
\begin{proofNestedLemma}
Consider a complete lattice $L=\langle \mathcal{L},\leq\rangle$ and suppose that $O$ is a  $\preceq^H_L$-monotonic operator.
Clearly, the set of post-fixpoints of $O$ is a partially ordered set. With upwards closedness, every chain of post-fixpoints has an upper bound (which is also a post-fixpoint), and thus, by the  Kuratowski-Zorn lemma \citep{kuratowski1922methode,zorn1935remark}, there is a maximal post-fixpoint. With Lemma \ref{lemma:maximal-post-fixpoint:also:fixpoint}, this maximal post-fixpoint is a fixpoint. 
\end{proofNestedLemma}
We have established thus that $\mathcal{O}_u(x,\cdot)$ admits a fixpoint (as $\mathcal{O}_u(x,\cdot)$ is a $\preceq^H_L$-monotonic operator for any (semi)-ndao $\mathcal{O}$ and any $x\in \mathcal{L}$). We can show, analogous to the proof of Proposition \ref{prop:stable:operator:is:subset:of:constructive:stable:operator}, that this means a well-founded sequence exists.
\end{appendixproof}

We now show that the c-stable operator is well-defined for all of the ndao's considered in this paper for choice programs $\mathcal{P}$ built up from choice atoms whose domain is finite. This is a sufficient condition: other conditions might also guarantee well-definedness. For most applications, this assumption seems warranted.
\begin{propositionAprep}
Let a choice program $\mathcal{P}$ s.t.\ for every $C_1\leftarrow C_2,\ldots,C_n\in \mathcal{P}$, $\dom(C_i)$ is finite for $i=1\ldots n$,
and ${\sf x}\in \{{\sf MR},{\sf LPST},\mathcal{U}\}$ and $x\subseteq y \subseteq \mathcal{A}$ be given.  Then $S^c(\ICc_\mathcal{P}^{\sf x})(x,y)\neq \emptyset$. Furthermore, $C^c(\ICc^{{\sf GZ},l}_\mathcal{P})(y)\neq \emptyset$.\footnote{Notice that, as $\ICc^{{\sf GZ},u}_\mathcal{P}$ is only $\preceq^H_L$-monotonic for consistent inputs, the complete operator for the upper-bound will not be well-defined as it starts from $\ICc_\mathcal{P}^{{\sf GZ},u}(x,\emptyset)$. This means that the ${\sf GZ}$-operator is only useful for total stable fixpoints.}
\end{propositionAprep}
\begin{appendixproof}
We first show $C^s(\ICc_\mathcal{P}^{{\sf x},l})(\cdot,y)\neq\emptyset$ for any $ y \subseteq \mathcal{A}$.
We have to show that $\ICc^{l,{\sf x}}_\mathcal{P}(\cdot,y)$ is downward closed
 $\ICc^{u,{\sf x}}_\mathcal{P}(x,\cdot)$ is upwards closed. 

We now show downward closedness. For this, let $\{x_\epsilon\}_{\epsilon\prec\alpha}$ be a descending chain of sets of atoms that are subsets of $y$ and are post-fixpoints of $\mathcal{IC}^{{\sf x},l}_\mathcal{P}(\cdot,y)$ for some $y\subseteq \mathcal{A}$,
and let $x=\bigcap \{x_\epsilon\}_{\epsilon\prec\alpha}$.  We show that $\mathcal{IC}^{{\sf x},l}_\mathcal{P}(x,y)\preceq^S_L \{x\}$. If the chain is finite, this is trivial. 
Suppose therefore that $\{x_\epsilon\}_{\epsilon\prec\alpha}$  is infinite.
\begin{description}
\item[${\sf x}={\sf MR}$]  Consider some $C\leftarrow C_1,\ldots,C_n\in \mathcal{P}$ s.t.\ for every $i=1,\ldots,n$, $y(C_i)={\sf T}$ and there is some $z'\subseteq x$ s.t.\ $z'(C_i)$. As $x=\bigcap \{x_\epsilon\}_{\epsilon\prec\alpha}$, also $z'\subseteq x_\epsilon$ for every $\epsilon\prec\alpha$. Thus, as $\mathcal{IC}^{{\sf x},l}_\mathcal{P}(x_\epsilon,y)\preceq^S_L \{x_\epsilon\}$, there is some $z_\epsilon\subseteq x_\epsilon$ s.t.\ $z_\epsilon\cap \dom(C)\in\sat(C)$ for every $\epsilon<\alpha$. As  $\{x_\epsilon\}_{\epsilon\prec\alpha}$ is a $\subseteq$-descending chain, and $\dom(C)$ is finite, there is some $\beta<\alpha$ s.t.\ for every $\epsilon,\epsilon'<\beta$, $\dom(C)\cap z_\epsilon=\dom(C)\cap z_\epsilon'\in\sat(C)$. But then also $\dom(C)\cap z_\epsilon'\subseteq x$, and thus $\mathcal{IC}^{{\sf x},l}_\mathcal{P}(x,y)\preceq^S_L \{x\}$.
\item[${\sf x}={\sf LPST}$] Consider some $C\leftarrow C_1,\ldots,C_n\in \mathcal{P}$ s.t.\ for every $i=1,\ldots,n$, and every $z\in [x,y]$, $z(C_i)={\sf T}$. As $x\subseteq x_\epsilon$ for every $\epsilon<\alpha$, $C\in\HDc_\mathcal{P}^{{\sf LPST},l}(x_\epsilon,y)$ (for every $\epsilon<\alpha$), and thus, as $\mathcal{IC}^{{\sf x},l}_\mathcal{P}(x_\epsilon,y)\preceq^S_L \{x_\epsilon\}$, there is some $z_\epsilon\subseteq x_\epsilon$ s.t.\ $z_\epsilon\cap \dom(C)\in\sat(C)$ for every $\epsilon<\alpha$. The rest of the proof is similar to the previous item.
\item[${\sf x}={\sf GZ}$]  Consider some $C\leftarrow C_1,\ldots,C_n\in \mathcal{P}$ s.t.\ for every $i=1,\ldots,n$, $x\cap\dom(C_i)=y\cap \dom(C_i)\in\sat(C_i)$. As $x\subseteq y$, for every $z\in [x,y]$, $z\cap\dom(C_i)=x\cap \dom(C_i)\in\sat(C_i)$. As $\mathcal{IC}^{{\sf x},l}_\mathcal{P}(x_\epsilon,y)\preceq^S_L \{x_\epsilon\}$, there is some $z_\epsilon\subseteq x_\epsilon$ s.t.\ $z_\epsilon\cap \dom(C)\in\sat(C)$ for every $\epsilon<\alpha$. The rest of the proof is similar to the previous item.
\item[${\sf x}=\mathcal{U}$]  We have to show that for every $z\in [x,y]$ there is some $z'\in \IC_\mathcal{P}(z)$ s.t.\ $z'\subseteq x$. As $z\in [x_\epsilon,y]$ for any $\epsilon<\alpha$, for every $\epsilon<\alpha$ there is some $z_\epsilon\subseteq x_\epsilon$ with $z_\epsilon\in \IC_\mathcal{P}(z)$. For every such $z_\epsilon$, it holds that for every $C\in\HD_\mathcal{P}(z)$, $z_\epsilon\cap \dom(C)\in\sat(C)$. As $\dom(C)$ is finite,  there is some $\beta<\alpha$ s.t.\ for every $\epsilon,\epsilon'<\beta$, $z_\epsilon\cap\dom(C)=z_{\epsilon'}\cap\dom(C)\in\sat(C)$ for every $\epsilon<\alpha$. But then also $z_\epsilon\cap\dom(C)\subseteq x$.
\end{description}

We now show upwards closedness of $\ICc^{\mathcal{U},u}_\mathcal{P}(x,\cdot)$  (for an arbitrary but fixed $x\subseteq \mathcal{A}$).
For this, let $\{y_\epsilon\}_{\epsilon<\alpha}$ be an ascending chain of sets of atoms that are supersets of $x$ and postfxipoints of $\ICc^{\mathcal{U},u}_\mathcal{P}(x,\cdot)$ and let $y=\bigcup \{y_\epsilon\}_{\epsilon<\alpha}$. We show that $\ICc^{\mathcal{U},u}_\mathcal{P}(x,y)\preceq^H_L \{y\}$. If the chain is finite, this is trivial. Suppose thus that $\{y_\epsilon\}_{\epsilon<\alpha}$ is infinite. We show that there is some $z\in [x,y]$ s.t.\ $y\in \IC_\mathcal{P}(y)$. Indeed, consider some $C\leftarrow C_1,\ldots,C_n$ s.t.\ $\dom(C_i)\cap y\in\sat(C_i)$ for every $i=1,\ldots,n$. 
As $y=\bigcup \{y_\epsilon\}_{\epsilon<\alpha}$, and $\dom(C_i)$ is finite (for every  $i=1,\ldots,n$), there is an $\epsilon<\alpha$ s.t.\ for every $\epsilon'\in [\epsilon,\alpha]$, $\dom(C_i)\cap y_{\epsilon'}\in \sat(C_i)$ (for $i=1,\ldots,n$). Thus, for every $\epsilon'\in [\epsilon,\alpha]$, $\dom(C)\cap y_{\epsilon'}\in \sat(C)$ (as $\{y_{\epsilon'}\}\preceq^H_L \ICc_\mathcal{P}^{\mathcal{U},u}(x,y_{\epsilon'})$).
Again in view of the finitude of $\dom(C)$, there is some $\epsilon<\alpha$ s.t.\ for every $\epsilon'\in [\epsilon,\alpha]$, $\dom(C)\cap y_{\epsilon}=\dom(C)\cap y_{\epsilon'}$, which implies that  $\dom(C)\cap y_{\epsilon}=\dom(C)\cap y$, and, since 
 $\dom(C)\cap y_{\epsilon}\in\sat(C)$, shows that $\dom(C)\cap y\in\sat(C)$.

That $C^c(\ICc^{{\sf x},u}_\mathcal{P})(x)\neq\emptyset$ follows from Lemma \ref{prop:h:monotonic:has:a:fp} and the fact that $\ICc_\mathcal{P}^{{\sf x},u}(x,\cdot)$ is $\preceq^H_L$-monotonic (Proposition \ref{prop:operators:are:ndaos}) for any $ x \subseteq \mathcal{A}$.
\end{appendixproof}

Upwards closedness is not guaranteed when allowing choice constructs with infinite domains:
\begin{example}
Consider the set of atoms $x=\{a_i\mid i\in \mathbb{N}\}$ and the choice atom $C=(A, \{A'\subseteq A\mid A'\mbox{ is infinite}\})$. We let $\mathcal{P}=\{C\leftarrow\}$. 
Let $x_j= A\setminus \{a_j\mid j\leq i\}$ for any $i\in\mathbb{N}$. Then $\{x_j\}_{j<\infty}$ is a $\subseteq$-descending chain of infinite sets. We see that for any $j<\infty$, $\IC_\mathcal{P}(x_j)=\{A'\subseteq A\mid A'\mbox{ is infinite}\}\preceq^S_L \{x_j\}$ as $x_j\in  \{A'\subseteq A\mid A'\mbox{ is infinite}\}$. However, $\bigcap_{j<\infty}\{x_j\}=\emptyset$, whereas $\IC_\mathcal{P}(\emptyset)=\{A'\subseteq A\mid A'\mbox{ is infinite}\}\preceq^S_L \{x_j\}$ and thus  $\IC_\mathcal{P}(\emptyset)\not\preceq^S_L \{\emptyset\}$.
\end{example}

\begin{remark}
Another property that stable fixpoints of both deterministic and non-deterministic operators have in common is that they are  $\leq_t$-minimal. As is to be expected and as can be seen from the constructive stable fixpoint $(\{p,q\},\{p,q\})$ in Example \ref{example:choice:rules:stable:2}, constructive stable fixpoints are not necessariy $\leq_t$-minimal, in contradistinction to minimiality-based stable fixpoints (Proposition 14 by \citet{heyninck2022non}).
\end{remark}

We now show a host of representation results: the ${\sf LPST}$-operator allows to generalize 
 the semantics of Liu et al \citep{liu2010logic} from two- to three-valued. The ${\sf GZ}$-operator allows to adapt the semantics of  \citet{gelfond2014vicious} (defined for disjunctive programs) to choice programs, and the semantics of \citet{marek2004set} from two- to three-valued.  

\begin{propositionAprep}\label{prop:characterisation:results}
Let a choice program $\mathcal{P}$ be given.
\begin{enumerate}
\item $x$ is a stable model according to \citet{liu2010logic} iff $(x,x)$ is a stable fixpoint of $\mathcal{IC}^{\sf LPST}_\mathcal{P}$. 
\item $x$ is a stable model according to \citet{marek2004set} iff $(x,x)$ is a stable fixpoint of $\mathcal{IC}^{\sf MR}_\mathcal{P}$. 
\item If $\mathcal{P}$ is a aggregate program then $x$ is a stable model according to \citet{gelfond2014vicious} iff 
 $x\in C^c(\mathcal{IC}^{{\sf GZ},l}_\mathcal{P})(x)$.
\end{enumerate}
\end{propositionAprep}
\begin{appendixproof}
For 1., we recall first some necessary preliminaries from \citep{liu2010logic}: 
\begin{definition}
Given some sets $x, y\subseteq \mathcal{A}$ and a choice atom $C$:
\begin{itemize}
\item  $ y\triangleright^{\mathrm{spt}}_x C$ iff $y$ is an $x$-trigger for $C$.
\item  $ \mathcal{P}^{\triangleright^{\mathrm{spt}}_x}(y)=\{C\leftarrow C_1,\ldots,C_n\mid y\triangleright^{\mathrm{spt}}_x C_i\mbox{ for }i=1\ldots n\}$,
\item  $T^{nd, \triangleright^{\mathrm{spt}}_x}_\mathcal{P}(y)=\{z\subseteq \bigcup_{C\in \mathcal{P}^{\triangleright^{\mathrm{spt}}_x}(y)}\dom(C)\mid z\cap \dom(C)\in\sat(C)\: \forall C\leftarrow \phi\in \mathcal{P}^{\triangleright^{\mathrm{spt}}_x}(y)\}$,
\item  A sequence  $\emptyset=x_1,\ldots,x_n=x$ is a self-justified  $\triangleright^{\mathrm{spt}}_x$-computation if $x_{i+1}\in T^{nd, \triangleright^{\mathrm{spt}}_x}_\mathcal{P}(x_i)$ for every $i\geq 1$, and $x\in T^{nd, \triangleright^{\mathrm{spt}}_x}_\mathcal{P}(x)$.
\end{itemize}
\end{definition}
We first show that $(\dagger)$: a sequence $\emptyset=x_1,\ldots,x_n=x$ is well-founded relative to $\ICc^{{\sf LPST},l}_\mathcal{P}(\cdot,x)$ iff it is a self-justified weak $ \triangleright^{\mathrm{spt}}_x$-computation. That this is the case is easily seen by observing that $\ICc_\mathcal{P}^{{\sf LPST},l}(x',x)=T^{nd, \triangleright^{\mathrm{spt}}_x}_\mathcal{P}(x')$ for any $x',x\subseteq \mathcal{A}$ (as $x'$ is an $x$-trigger for $C$ iff for every $z\in[x',x]$, $z(C)={\sf T}$). 
We now show the main claim.
Suppose for the $\Rightarrow$-direction that $x$ is a stable model according to \citep{liu2010logic}. With Proposition 12 from \citet{liu2010logic}, there is a self-justified   $\triangleright^{\mathrm{spt}}_x$-computation for $x$. With $\dagger$, there is a well-founded sequence  relative to $\ICc^{{\sf LPST},l}_\mathcal{P}(\cdot,x)$ for $x$. Thus, $x\in  S^c(\mathcal{IC}^{\sf LPST}_\mathcal{P})(x,x)$. Suppose now for the $\Leftarrow$-direction that  $x\in  S^c(\mathcal{IC}^{\sf LPST}_\mathcal{P})(x,x)$, i.e.\ there is a well-founded sequence  relative to $\ICc^{{\sf LPST},l}_\mathcal{P}(\cdot,x)$ for $x$. With $\dagger$,  this sequence  is a self-justified weak $ \triangleright^{\mathrm{spt}}_x$-computation. With Proposition 11 by \citet{liu2010logic}, this is a founded computation. By Definition 5 by \citet{liu2010logic}, this means $x$ is an answer set according to \citet{liu2010logic}.

2. We first recall the relevant definitions \citep{marek2004set} (adapted to our notation):
\newcommand{\nss}{\mathrm{NSS}}
Given a choice atom $C=(\dom,\sat)$, and a set of atoms $x$, $\overline{C}_y=\{z\subseteq y\mid \exists z'\in \sat(C)\mbox{ and }z\subseteq z'\}$. A perhaps helpful alternative description is: $\overline{C}_y=\bigcup_{z\in \sat(C)}^{z\subseteq y}[z,y]$. 
Given a choice program $\mathcal{P}$ and a set of atoms $x$, $\nss(\mathcal{P},y)$ is obtained by:
\begin{enumerate}
\item Removing any $C\leftarrow C_1,\ldots,C_n$ s.t.\ $y(C_i)={\sf F}$ for some $i=1,\ldots,n$,
\item For every rule $r=C\leftarrow C_1,\ldots,C_n$ not removed in the first step, replace $r$ by $\{ \alpha \leftarrow \overline{C_1}_x,\ldots,\overline{C_n}_x\mid \alpha\in\dom(C)\}$.
\end{enumerate}
 \citet{marek2004set} show that for any choice program $\mathcal{P}$, $\nss(\mathcal{P},y)$ has a unique least model obtained by iteratively applying the one-step provability operator  $T_{\nss(\mathcal{P},y)}(x)=\{ \alpha\mid \alpha \leftarrow \overline{C_1}_x,\ldots,\overline{C_n}_x\in \nss(\mathcal{P},y)\mbox{ and }\forall  i=1\ldots n: x\cap \dom( \overline{C_i}_x)\in\sat(\overline{C_i}_x)\}$. Call this least model $N_{\mathcal{P},y}$. Then $y$ is a ${\sf MR}$-stable model of $\mathcal{P}$ iff $y$ is a model of $\mathcal{P}$ and $y=N_{\mathcal{P},y}$.
 We show that ($\dagger$): $T_{\nss(\mathcal{P},x)}(z)\in \ICc^{{\sf MR},l}_\mathcal{P}(z,x)$ for any $z\subseteq x$. For this, consider some $C\leftarrow C_1,\ldots,C_n\in \mathcal{P}$. We distinguish two cases:
 \begin{enumerate}
\item %
$x(C_i)={\sf F}$ for some $i=1,\ldots,n$. Then $C\not\in \HDc^{{\sf MR},l}_\mathcal{P}(z,x)$ for any $z\subseteq x$ and thus we are done.
\item %
$x(C_i)={\sf T}$ for every  $i=1,\ldots,n$.  We distinguish two cases:
\begin{enumerate}
\item 
Suppose first that there is no $z\subseteq x$ s.t.\ $z(C_i)={\sf T}$ for some $i=1,\ldots,n$. Then $C\not\in \HDc^{{\sf MR},l}_\mathcal{P}(z,x)$ and thus we are done. 
\item 
Suppose now that there is a $z'\subseteq z$ s.t.\ $z'(C_i)={\sf T}$ for every $i=1,\ldots,n$.  Then $C\in \HDc^{{\sf MR},l}_\mathcal{P}(z,x)$.
Also, for every $i=1,\ldots,n$, $z\in \sat(\overline{C_i}_x)$ and, as $\alpha\leftarrow \overline{C_1}_x,\ldots,  \overline{C_n}_x\in \nss(\mathcal{P},x)$ for every $\alpha\in \dom(C)\cap x$, $\dom(C)\cap x\subseteq  T_{\nss(\mathcal{P},x)}(z)$. As $x$ is a model of $\mathcal{P}$ and $x(C_i)={\sf T}$ for every $i=1,\ldots,n$, $x\cap\dom(C)\in\sat(C)$. 
\end{enumerate}
 \end{enumerate} 
We now show that $x\in S^c(\ICc^{{\sf MR},l}_\mathcal{P})(x,x)$ iff $x$ is a ${\sf MR}$-stable model of $\mathcal{P}$.
For the $\Rightarrow$-direction, observe first that as $x$ is a fixpoint of $\ICc^{{\sf MR},l}_\mathcal{P}$, it is, with Proposition \ref{prop:supported}, a model of $\mathcal{P}$. With $\dagger$, $T_{\nss(\mathcal{P},x)}(z)\in \ICc^{{\sf MR},l}_\mathcal{P}(z,x)$ for any $z\subseteq x$, which implies
(with a simple induction on the well-founded sequence $\emptyset,\ldots,x$) that $x=N_{\mathcal{P},x}$.
For the $\Leftarrow$-direction, suppose that $x$ is an ${\sf MR}$-stable model. Notice that as $x=N_{\mathcal{P},x}$, there is a sequence $\emptyset=x_0, x_1,\ldots,x_n=x$ s.t.\ $x_{i+1}= T_{\nss(\mathcal{P},x)}(x_i)$ for any $i=0,\ldots,n$. With $\dagger$, this is a well-founded sequence relative to $\ICc_\mathcal{P}^{{\sf MR},l}(\cdot,x)$. 

3.\ follows from the fact that for choice programs with only atoms in the head, $\ICc^{{\sf GZ},l}$ coincides with $\mathcal{I}^{\sf GZ}_{\mathcal{P},x}(y)$, used by \citet{alviano2023aggregate} to characterize the semantics of \citet{gelfond2014vicious}.
\end{appendixproof}
Furthermore,  for normal logic programs, all operators besides the ultimate coincide, and the stable fixpoints respectively partial stable fixpoint coincide with the stable respectively partial stable models (and similarly for the well-founded semantics) (recall Remark \ref{remark:semanticsNLPs:are:preserved}). Thus, the semantics studied in this paper do not only generalize existing semantics for choice or aggregate programs, but also the well-known semantics for normal logic programs.

\section{Groundedness}\label{sec:postulates}
 We  introduce several postulates to facilitate a comparison between semantics for choice programs (thus solving an open question in the literature \citet{alviano2023aggregate}) formalizing in different ways the idea of \emph{groundedness}. Furthermore, we show that for every notion of groundedness, there exist examples which have been argued in the literature to be counter-intuitive.
Intuitively, the idea behind groundedness is that models should be derivable \emph{from the ground up}, i.e.\ they should be supported by non-cyclic arguments. For choice programs,  what constitutes a cycle becomes less clear:
\begin{example}[\citep{liu2010logic}]\label{ex:illustration:of:semantics}
Consider the program $\mathcal{P}=\{ \{p,q\}=2\leftarrow \{p,q\}\neq 1\}$.
There are two candidates for stable models: $\emptyset$ and $\{p,q\}$ (as the only satisfier of the head of the only rule is $\{p,q\}$), which we discuss:\\
\noindent(1) If we choose $\emptyset$, then we see that the only rule in the program is applicable but not applied.\\
\noindent(2)
 If we choose $\{p,q\}$, we could justify this intuitively by the sequence $\emptyset,\{p,q\}$: at the first step, $\emptyset$ makes $\{p;q\}\neq 1$ true and thus we derive $\{p,q\}$. At the second step, however, we can only justify our choice by a self-supporting justification: $\{p,q\}$ is true since it is derivable using the head of the rule $ \{p,q\}=2\leftarrow  \{p;q\}\neq 1$ and since $\{p,q\}(\{p;q\}\neq 1)={\sf T}$.

 We observe that $\{p,q\}$ is stable fixpoint for the $\mathcal{U}$- and ${\sf MR}$-operator, but not for the ${\sf LPST}$- and ${\sf GZ}$-operators. 
\end{example}
The benefit of our operator-based framework is that we do not have to choose for a single ``best'' semantics: the choices outlined above will be obtainable by using different operators. 
We now carry forward our study of self-support on a more general level by introducing three postulates of decreasing strength.
The first notion (inspired by \citet{gelfond2014vicious}) requires that we can find a stratification of the program (i.e.\ an assignment $\kappa$ of natural numbers representing levels) s.t.\ the entire domain of the head of a rule is strictly higher stratified then the domain of the choice constructs in the body.
\begin{definition}
 A set $x$ is \emph{d(omain)-grounded} (for $\mathcal{P}$) if there is some $\kappa:x\rightarrow\mathbb{N}$ s.t.\ every $a\in x$ there is some $r=C\leftarrow C_1,\ldots,C_n\in \mathcal{P}$ s.t.\ $a\in\dom(C)$, $\kappa(a)>\max \{\kappa(b)\mid b\in \bigcup_{i=1}^n\dom(C_i)\}$. 
\end{definition}
As already observed by \citet{alviano2023aggregate}, this requirement might be overly strong:
\begin{example}\label{example:domain:grounded}
Consider  $\mathcal{P}=\{b\leftarrow 1 \{a,b\}; a\leftarrow\}$. Then one might expect $\{a,b\}$ to be a good candidate for a stable model. Indeed, $a$ is a fact and allows to support $b$. However, $\{a,b\}$ is not d-grounded, as there is no $\kappa:\{a,b\}\rightarrow \mathbb{N}$ s.t.\ $\kappa(b)>\max \{\kappa(a),\kappa(b)\}$. 
\end{example}
 The following weaker notion (inspired in name and idea by  \citet{liu2010logic}) requires that at some point in a sequence, there is support for the body of a rule, and this support persists in every following step:
\begin{definition}
(1) A set $x$ is an \emph{$y$-trigger} for $C\leftarrow C_1,\ldots,C_n$ if for every $z\in [x,y]$, $z(C_i)={\sf T}$ for every $i=1,\ldots,n$.
(2) A set $x$ is \emph{s(trongly)-grounded} (for $\mathcal{P}$) if there is some $\kappa:x\rightarrow\mathbb{N}$ s.t.\ every $a\in x$, there is some $r=C\leftarrow C_1,\ldots,C_n\in \mathcal{P}$ s.t.\ $a\in\dom(C)$ and there is an $x$-trigger $z$ for $r$ s.t.\ $\kappa(a)>max\{\kappa(b)\mid b\in z\}$.
\end{definition}
Intuitively, $x$ is s-grounded if for every atom $a\in x$, we can find a rule that that has an $x$-trigger in a strictly lower level. 

D-grounded sets are s-grounded, but not vice-versa:
\begin{example}
Consider $\mathcal{P}$ as in Example \ref{example:domain:grounded}. 
We see that $\{a,b\}$ is strongly grounded. Indeed, if we take $\kappa(a)=0$ and $\kappa(b)=1$, we see that 
$b\leftarrow 1 \{a,b\}$ has a $\kappa$-lower $\{a,b\}$-trigger in $\{a\}$ as $\{a\}(1 \{a,b\})=\{a,b\}(1 \{a,b\})={\sf T}$. 
\end{example}

The following example, first introduced by \citet{DBLP:conf/jelia/Alviano019}, shows that in some circumstances, s-groundedness might be overly strong:
\begin{example}\label{example:chain:answer:sets}
Consider $\mathcal{P}=\{a\leftarrow \{a,b\}\neq 1; b\leftarrow \{a,b\}\neq 1\}$. One might expect the set $\{a,b\}$ to be acceptable. Suppose it is s-grounded. Then there is an $\{a,b\}$-trigger $y$ for $a$ s.t.\ $y\subseteq \{b\}$. This is impossible as $\{b\}\in [\emptyset,\{a,b\}]$ and $\{b\}\in [\{b\},\{a,b\}]$ and $\{b\}\cap \dom(\{a,b\}\neq 1)\not\in \sat(\{a,b\})$. Thus, $\{a,b\}$ cannot be s-grounded.
\end{example}

We thus introduce an even weaker notion of groundedness which we call \emph{antecedent groundedness}, which requires that for every accepted atom, we can find a rule that is activated by a lower $\kappa$-level:
\begin{definition}
A set $x$ is \emph{a(ntecedent)-grounded} if there is some $\kappa:x\rightarrow\mathbb{N}$ s.t.\ every $a\in x$, there is some $r=C\leftarrow C_1,\ldots,C_n\in \mathcal{P}$ s.t.\ $a\in\dom(C)$ and  for every $i=1,\ldots,n$ there is some $z\subseteq \{b\mid \kappa(b)< \kappa(a)\}$ s.t.\ $z(C_i)={\sf T}$ .
\end{definition}
S-grounded sets are a-grounded, but not vice-versa:
\begin{example}
Consider again $\mathcal{P}$ from Example \ref{ex:illustration:of:semantics}. We first observe that $\{p,q\}$ is antecedent grounded as $\emptyset(\{p,q\}\neq 1)={\sf T}$ and thus we have found our antecedent justification for $\{p,q\}$. However, $\emptyset$ is \emph{not} a $\{p,q\}$-trigger, as $\{p\}\in [\emptyset,\{p,q\}]$ and $\{p\}(\{p,q\}\neq 1)={\sf F}$. Likewise, it can be seen that $\{a,b\}$ in Example \ref{example:chain:answer:sets} is a-grounded.
\end{example}

The reader might think that a-groundedness is the most suitable notion of groundedness, as it avoids the behavior (deemed problematic by \citet{DBLP:conf/jelia/Alviano019} in Example \ref{example:chain:answer:sets}). However,  \citet{liu2010logic} argued that, for $\mathcal{P}$ from Example \ref{ex:illustration:of:semantics}, $\{p,q\}$ is not a feasible candidate for an answer set. Altogether, we see that for every notion of groundedness, one can find examples deemed as counter-intuitive in the literature.

Notice that all notions of groundedness generalize a well-known notion of groundedness proposed for normal logic programs \citep{erdem2003tight}:
\begin{propositionAprep}\label{prop:a:grounded:then:redem:grounded}
Let $\mathcal{P}$ a normal logic program $\mathcal{P}$. Then $x$ is a-grounded $\mathcal{P}$ then $x$ is grounded according to \citet{erdem2003tight}, namely: there is a ranking $\kappa:x\rightarrow\mathbb{N}$ s.t.\ for every $a\in x$, there is some $a\leftarrow a_1,\ldots,a_n, \lnot b_1,\ldots,\lnot b_m\in\mathcal{P}$ s.t.\ for every $i=1,\ldots,n$, $\kappa(a)>\kappa(a_i)$.
\end{propositionAprep}
\begin{appendixproof}
Suppose that $x$ is a-grounded for $\mathcal{P}$, i.e.\ there is some $\kappa:x\rightarrow\mathbb{N}$ s.t.\ for every $a\in x$, there is some $r=a\leftarrow \bigwedge_{i=1}^n a_i\land \bigwedge_{i=1}^m \lnot b_i$ s.t.\ for every $i=1,\ldots,n$, there is some $z\subseteq \{b\mid \kappa(b)<\kappa(a)\}$ with $z(a_i)={\sf T}$. As $z(a_i)={\sf T}$ iff $a_i\in z$, this means that $\kappa(a)>\kappa(a_i)$ for every $i=1,\ldots,n$.
\end{appendixproof}
\begin{toappendix}
\begin{remark}
Notice that the other direction of Proposition \ref{prop:a:grounded:then:redem:grounded} does not necessarily hold: take e.g.\ $\mathcal{P}=\{b\leftarrow\lnot c., c\leftarrow \lnot b.\}$. Then $\{b,c\}$ is grounded according to \citet{erdem2003tight} (trivially, as there are no rules with positive bodies) but not a-grounded.
\end{remark}
\end{toappendix}

The $\mathcal{U}$-operator does not satisfy any notion of groundedness, whereas all other operators give rise to a-grounded stable models, but only the ${\sf LPST}$-operator and ${\sf GZ}$-operator give rise to s-grounded stable models and only the ${\sf GZ}$-operator gives rise to d-grounded stable models:
\begin{propositionAprep}
Let $\mathcal{P}$ be a choice program and $\dagger\in \{{\sf GZ},{\sf LPST},{\sf MR}\}$. Then if $(x,y)\in S^c(\mathcal{IC}_\mathcal{P}^\dagger)(x,y)$:
\begin{enumerate}
\item $x$ and $y$ are a-grounded (for $\mathcal{P}$), and
\item $x$ is s-grounded (for $\mathcal{P}$) if $\dagger\in\{{\sf GZ},{\sf LPST}\}$.
\item $x$ and $y$ are d-grounded (for $\mathcal{P}$) if $\dagger={\sf GZ}$.
\end{enumerate}
There are programs $\mathcal{P}$ s.t.\ $(x,y)\in S^c(\mathcal{IC}_\mathcal{P}^\mathcal{U})(x,y)$ yet $x$ and $y$ are not antecedent grounded.
\end{propositionAprep}
\begin{appendixproof}
Let $\mathcal{P}$ be a choice program and $\dagger\in \{{\sf GZ},{\sf LPST},{\sf MR}\}$. 

Ad 1.: 
We first consider $ C^c(\mathcal{IC}^{l,{\sf MR}}_\mathcal{P})$.
Suppose that $x\in C^c(\mathcal{IC}^{l,{\sf MR}}_\mathcal{P})(y)$, i.e.\ there is a well-founded sequence $\emptyset=x_0,\ldots,x_n,x$ relative to $\ICc^{{\sf MR},l}_\mathcal{P}(\cdot,y)$.
We let $\kappa(a)=\min \{i\mid a\in x_i\}$. 
 Consider some $a\in x$ and let $i$ be the ordinal s.t.\ $a\in x_{i+1}\setminus x_i$. Then there is some $C\leftarrow C_1,\ldots,C_n\in\mathcal{P}$ s.t.\ $a\in  \dom(C)\cap x_{i+1}$ and for every $j=1\ldots n$, $y(C_j)={\sf T}$ and there is some $w\subseteq x_{i}$ s.t.\ $w(C_j)={\sf T}$. But then $w(C_j)={\sf T}$ and $\kappa(b)\leq i < \kappa(a)=i+1$.\\ 
 We now consider $ C^c(\mathcal{IC}^{{\sf GZ},u}_\mathcal{P})=C^c(\mathcal{IC}^{{\sf LPST},u}_\mathcal{P})=C^c(\mathcal{IC}^{\mathcal{U},u}_\mathcal{P})$.  
Suppose that  $y\in C^c(\mathcal{IC}^{u,{\sf MR}}_\mathcal{P})(x)$, i.e.\ there is a well-founded sequence $\emptyset=y_0,\ldots,y_n,y$ relative to $\ICc^{\mathcal{U},u}_\mathcal{P}(x,\cdot)$.
We let $\kappa(a)=\min \{i\mid a\in y_i\}$. 
Consider some $a\in y$ and let $i$ be the ordinal s.t.\ $a\in y_{i+1}\setminus y_i$. Then there is some $z\in [x,y_{i}]$ and some $C\leftarrow C_1,\ldots,C_n$ s.t.\ $z(C_j)={\sf T}$ for every $j=1,\ldots,n$ and $a\in\dom(C)$. As $z\subseteq y_i$, $z\subseteq \{a\in\mathcal{A}\mid \kappa(z)\leq i\}$, whereas $\kappa(a)=i+1$. Thus, we have found our justification for $a$.

 Ad 2.: Suppose that $x$ is a constructive stable fixpoint of $\mathcal{IC}^{\sf LPST}_\mathcal{P}$, i.e.\ there is a well-founded sequence relative to $\ICc^{{\sf LPST},l}_\mathcal{P}(\cdot,y)$ $\emptyset=x_0,\ldots,x_n,x$. We let $\kappa(a)=\min \{i\mid a\in x_i\}$.  Consider some $a\in x$ and let $i$ be the ordinal s.t.\ $a\in x_{i+1}\setminus x_i$.
Then there is some $r=C\leftarrow C_1,\ldots,C_n\in\mathcal{P}$ s.t.\ $a\in  \dom(C)\cap  x_{i+1}$ and for every $z\in [x_i,y]$ and every $j=1,\ldots,n$, $z(C_j)={\sf T}$. But then clearly $x_i$ is an $y$-trigger for $r$ and for every $b\in x_i$, $\kappa(b)\leq i<\kappa(a)=i+1$.

Ad 3.: suppose that $x$ is a constructive stable fixpoint of $\mathcal{IC}^{\sf GZ}_\mathcal{P}$,  i.e.\ there is a well-founded sequence relative to $\ICc^{{\sf GZ},l}_\mathcal{P}(\cdot,y)$ $\emptyset=x_0,\ldots,x_n,x$.  We let $\kappa(a)=\min \{i\mid a\in x_i\}$. Consider now some $a\in x$ and let $i$ be the ordinal s.t.\ $a\in x_{i+1}\setminus x_i$. This means that $a\in\dom(C)$ for some $C\in \HDc^{{\sf GZ},l}_\mathcal{P}(x_i,x)$. Suppose towards a contradiction that for every $C\leftarrow C_1,\ldots,C_n\in \mathcal{P}$ with $a\in \dom(C)$, there is some $i=1,\ldots,n$ s.t.\ for some $b\in\dom(C_i)$, $\kappa(b)\geq \kappa(a)$. For any such $b$, $b\not\in x_i$, which implies that $b\not\in\dom(C_i)\cap x_i$ yet $b\in \dom(C_i)\cap x$, i.e.\ $\dom(C_i)\cap x_i\neq \dom(C_i)\cap x$, contradiction to $C\in \HDc^{{\sf GZ},l}_\mathcal{P}(x_i,x)$.

Regarding the ultimate operator, consider $\mathcal{P}=\{\{p,q\}=2\leftarrow \{p,q\}=2\}$. Then we see that $\ICc^{\mathcal{U}}_\mathcal{P}(\emptyset,\{p,q\})=(\bigcup_{z\subseteq \{p,q\}}\IC_\mathcal{P}(z),\bigcup_{z\subseteq \{p,q\}}\IC_\mathcal{P}(z))=(\{\emptyset,\{p,q\}\}, \{\emptyset,\{p,q\}\})$. Thus, the sequence $\emptyset,\{p,q\}$ is well-founded relative to $\ICc^{\mathcal{U},l}_\mathcal{P}(\cdot,\{p,q\})$. However, it is clearly not antecedent grounded as  we cannot justify $\{p,q\}$ in view of $\emptyset$. 
\end{appendixproof}

Summing up, we see that the different semantics satisfy different notions of groundedness, and that no notion of groundedness is uncontested. It seems that it depends on the application at hand which notion of groundedness is the most suitable.
The results of this section are summarized in Table \ref{table:postulates}.
\begin{table*}
\centering
\begin{tabular}{l|lll}
\toprule Operator  &d-ground.& s-ground.& a-ground. \\ \midrule 
   ${\sf GZ}$& \cellcolor{green!25}$\vee$ & \cellcolor{green!25}$\vee$ &\cellcolor{green!25}$\vee$ \\
 ${\sf LPST}$& \cellcolor{red!25}$\times$ & \cellcolor{green!25}$\vee$ & \cellcolor{green!25}$\vee$\\
  ${\sf MR}$& \cellcolor{red!25}$\times$ &  \cellcolor{red!25}$\times$ & \cellcolor{green!25}$\vee$\\
  $\mathcal{U}$& \cellcolor{red!25}$\times$ & \cellcolor{red!25}$\times$& \cellcolor{red!25}$\times$\\ \midrule \midrule 
  Example of violation &   Ex.\ \ref{example:domain:grounded} & Ex.\ \ref{example:chain:answer:sets} & Ex.\ \ref{ex:illustration:of:semantics}\\
Counter-intuitive according to &  \citep{alviano2023aggregate} & \citep{DBLP:conf/jelia/Alviano019}  &\citep{liu2010logic}\\ \bottomrule 
\end{tabular}
\caption{Results on Postulates}\label{table:postulates}
\end{table*}

\section{Disjunctions are Choice Constructs}\label{sec:disjunction}
Our study allows us to give a principled account of the relation between stable semantics for disjunctive logic programs (DLPs) and choice programs. Indeed, in this section, we show that for DLPs \citep{heyninck2022non,DBLP:journals/corr/abs-2305-10846} are a special case of the operator for choice programs. This means that all semantics obtained on the basis of these operators coincide, thus explaining the difference between semantics for stable models of disjunctive logic programs and choice logic programs in terms of the choice of stable operator (minimality-based versus constructive). In the rest of this section, we show this claim in more detail, first providing the necessary background on DLPs and then showing DLP-operators are a special case of the operators studied in this paper.

\paragraph*{Preliminaries on disjunctive logic programs}
We first recall some necessary preliminaries on disjunctive logic programming. For simplicity, we restrict attention to disjunctive logic programs whose bodies consist solely of literals, leaving the generalization of these results to stronger languages for future work. 

In what follows we consider a propositional\footnote{We restrict ourselves to the propositional case.} language ${\mathfrak L}$, 
whose atomic formulas are denoted by $p,q,r$ (possibly indexed), and that contains the propositional constants ${\sf T}$ (representing truth), ${\sf F}$ (falsity), 
${\sf U}$ (unknown), and ${\sf C}$ (contradictory information). The connectives in  ${\mathfrak L}$ include negation $\neg$, conjunction $\wedge$ and disjunction $\vee$. Formulas are denoted by $\phi$, $\psi$, $\delta$ (again, possibly indexed). A (propositional) {\em disjunctively normal logic program\/} $\mathcal{P}$ in ${\mathfrak L}$ (a dlp in short) is a finite set of rules of the form 
             $\bigvee_{i=1}^n p_i~\leftarrow~\bigwedge_{i=1}^n a_i\land \bigwedge_{i=1}^m \lnot b_i$,  where the head $\bigvee_{i=1}^n p_i$ is a non-empty disjunction of atoms, and the body $\psi$
             is a formula in ${\mathfrak L}$.  We furthermore recall that a formula of the form $\bigwedge_{i=1}^n a_i\land \bigwedge_{i=1}^m \lnot b_i$ can be evaluated over the bilattice {\sf FOUR} relative to $(x,y)$ by assuming the involution $-$ defined by $-{\sf T}={\sf F}$, $-{\sf F}={\sf T}$, $-{\sf U}={\sf U}$ and $-{\sf C}={\sf C}$,
 and by defining truth assignments to complex formulas recursively as follows:
 
    \begin{itemize}\item $(x,y)({p})=
\begin{cases}
      {\sf T} & \text{ if } {p} \in x \text{ and } {p} \in y,  \\
      {\sf U} & \text{ if } {p} \not\in x \text{ and }{p} \in y,  \\
      {\sf F} & \text{ if } {p} \not\in x \text{ and } {p} \not\in y,  \\
      {\sf C} & \text{ if } {p} \in x \text{ and } {p} \not\in y. 
\end{cases}$ \smallskip
\item $(x,y)(\lnot \phi)=- (x,y)(\phi)$,  
\item $(x,y)(\psi \land \phi)=\sqcap_{\leq_t}\{(x,y)(\phi),(x,y)(\psi)\}$, 
\item $(x,y)(\psi \lor \phi)= \sqcup_{\leq_t}\{(x,y)(\phi),(x,y)(\psi)\}$. 
\end{itemize}
Notice that this is equivalent to the evaluations introduced in Section \ref{sec:supported:models}.

We recall the following ndao $\ICc^d_\mathcal{P}$ introduced by \citep{heyninck2022non} and defined as follows (given a dlp $\mathcal{P}$ and an interpretation $(x,y)$):
\begin{itemize}
\item  $\HRc^{d,l}_\mathcal{P}(x,y) = \{ \Delta \mid \bigvee\!\Delta \leftarrow \phi\in \mathcal{P}, (x,y)(\phi)\geq_t {\sf C}\}$, 
\item   $\HRc^{d,u}_\mathcal{P}(x,y) = \{ \Delta \mid \bigvee\!\Delta \leftarrow \phi\in \mathcal{P}, (x,y)(\phi)\geq_t {\sf U}\}$, 
 \item   $\ICc^{d,\dagger}_\mathcal{P}(x,y)=\{x_1\subseteq \bigcup\HRc^{d,\dagger}_\mathcal{P}(x,y) \mid \forall \Delta\in \HRc^{d,\dagger}_\mathcal{P}(x,y), \ x_1 \cap \Delta \neq \emptyset \}$ (for $\dagger\in \{l,u\}$),
\item   $\ICc^d_\mathcal{P}(x,y)=(\ICc^{d,l}_\mathcal{P}(x,y), \ICc^{d,u}_\mathcal{P}(x,y))$. 
\end{itemize}

The operator $\ICc^d_\mathcal{P}$ faithfully represents the semantics of dlps:
In general, total stable fixpoints of $\mathcal{P}$ correspond to stable models of $\mathcal{P}$ \citep{gelfond1991classical}, and weakly supported models of $\mathcal{P}$ \citep{brass1995characterizations} correspond to fixpoints of $\ICc^d_\mathcal{P}$ \citep{heyninck2022non}.

\paragraph*{Disjunctions as Choice Constructs}
We now show how the operator $\ICc^d_\mathcal{P}$ can be seen as a special case of the operators for choice constructs. In more detail, 
for a DLP $\mathcal{P}$, we define ${\tt D2C}(\mathcal{P})=\{1\Delta\leftarrow \phi\mid \bigvee\Delta\leftarrow \phi\in\mathcal{P}\}$. E.g.\ ${\tt D2C}(\{p\lor q\leftarrow\})= \{1\{p,q\}\leftarrow\}$.  In other words, we replace every disjunction by the choice atom that requires at least one element of $\Delta$ is true. 
We will show here that the operator defined for disjunctive logic programs \citep{heyninck2022non} then coincides with the operator $\ICc^{\sf x}_{{\tt D2C}(\mathcal{P})}$ (for ${\sf x}={\sf MR},{\sf LPST}$). 
This implies that all AFT-based semantics coincide for DLPs and their conversion into choice rules. We can now point in a very exact way to the difference between DLPs and choice programs: it lies not in how the constructs of disjunction and choice atoms are treated (i.e.\ when they should be made true or false), but rather in how the stable semantics is defined: for disjunctions, typically (e.g.\ in the most popular solvers \citep{gebser2016theory,eiter2012conflict}), a minimality-based stable operator is used, whereas for choice constructs, the c-stable operator is more apt. Thus, disjunctive and choice programs use the same (approximation) operators, but differ in how the corresponding stable operator is constructed.

We now show that the operator $\ICc^d_\mathcal{P}$ is a special case of the ${\sf MR}$- and ${\sf LPST}$-operators from this paper, when applied to the choice program ${\tt D2C}(\mathcal{P})$.
\begin{propositionAprep}
For any disjunctive normal logic program $\mathcal{P}$, $\ICc_\mathcal{P}=\ICc^{\sf MR}_{{\tt D2C}(\mathcal{P})}=\ICc^{\sf LPST}_{{\tt D2C}(\mathcal{P})}$.
\end{propositionAprep}
\begin{appendixproof}
We first notice that, as ${\tt D2C}(\mathcal{P})$ does not change the bodies of rules in $\mathcal{P}$, if $\mathcal{P}$ is a disjunctively normal logic program, ${\tt D2C}(\mathcal{P})$ is a normal choice program. This means that,  with Proposition \ref{prop:normal:choice:MR:LPST:same}, $\ICc^{\sf MR}_{{\tt D2C}(\mathcal{P})}=\ICc^{\sf LPST}_{{\tt D2C}(\mathcal{P})}$.

Furthermore, it can be straightforwardly shown that for any disjunctively normal logic program, $\HRc^{d,\dagger}_\mathcal{P}(x,y)=\HRc^{{\sf MR},\dagger}_\mathcal{P}(x,y)$.
Finally, it suffices to observe that $z\cap \dom(1\leq\Delta)\in \sat(1\leq\Delta)$ iff $\Delta\cap z\neq \emptyset$ for any $z\subseteq \mathcal{A}$, which means that $z\in \ICc_\mathcal{P}(x,y)$ iff $z\in\ICc^{\sf MR}_{{\tt D2C}(\mathcal{P})}(x,y)$ for any $x,y\subseteq \mathcal{A}$.
\end{appendixproof}
From this, we immediately obtain that all the major semantics for disjunctive logic programming are special cases of the semantics introduced in this paper. Namely, weakly supported models of $\mathcal{P}$ coincide with fixpoints of $\ICc^{\sf MR}_{{\tt D2C}(\mathcal{P})}$ whereas minimality-based stable fixpoints of $\mathcal{P}$ coincide with stable models of $\mathcal{P}$. Since, the c-stable semantics  does not enforce minimality, it will in general \emph{not} correspond to the stable models semantics:
\begin{example}
Let $\mathcal{P}=\{p\lor q\leftarrow\}$. Then $(\{p,q\},\{p,q\})$ is a c-stable fixpoint of $\ICc^{\sf MR}_{{\tt D2C}(\mathcal{P})}$ but it is not a stable model of $\mathcal{P}$ according to  \citet{gelfond1991classical}.
\end{example}

This also gives an answer to the question of how to combine disjunctions and choice constructs in logic programs: a choice as to which stable semantics are used has to be made: either preserve the minimality of answer sets as in DLPs and loses some reasonable models, or one gives up the minimality requirement by using the constructive stable semantics. In this context, it is perhaps interesting to note that the constructive stable semantics still coincides with the standard stable semantics for normal logic programs. In that case, all stable models are minimal. On the other hand, we can now also use minimality-based stable semantics on more complicated choice construct than simple disjunctions.

\section{Related Work}\label{sec:rel:work}
To the best of our knowledge, this is the first application of AFT to the semantics of choice programs. 
We have shown how major semantics for choice programs \citep{marek2004set,liu2010logic} can be characterized in our framework. Other well-known semantics for (disjunctive) aggregate programs are those by \citet{faber2004recursive,DBLP:conf/jelia/Alviano019,ferraris2011logic}, which are the ones used in the solvers DLV respectively clingo \citep{ferraris2011logic,alviano2017asp}.

Another semantics, the FLP-semantics \citep{faber2004recursive}, were originally not defined for programs with choice constructs in the head, but were generalized to allow choice constructs in the head by \citet{eiter2023explaining}. Nevertheless, these semantics only allow for two-valued stable models, and were shown  to differ from the semantics by \citep{gelfond2014vicious}, \citep{liu2010logic} and \citep{marek2004set} already for aggregate programs \citep{alviano2023aggregate}, which means, in view of Proposition \ref{prop:characterisation:results}, that the stable semantics induced by the ${\sf LPST}$-, ${\sf MR}$- and ${\sf GZ}$-operators also differ from the FLP-semantics. This comparison also holds for the semantics by \citet{ferraris2011logic} as the latter coincide with the FLP-semantics for aggregates with positive atoms \citep{alviano2023aggregate}. 

Another line of work that is relevant for this paper is the application of AFT to logic programs with aggregates.
\citet{pelov2007well} introduce several approximation operators for aggregate programs, whereas \cite{pelov2007well} introduces operator-based semantics for disjunctive aggregate programs. These semantics were generalized by \cite{DBLP:journals/corr/abs-2305-10846} in the framework of non-deterministic AFT. Generalizing the operator by   \cite{pelov2007well} to choice programs is left for future work. An overview of semantics that are (non-)representable in the deterministic AFT-framework is given by \citet{vanbesien2022analyzing}. This work severed as an important foundation of our paper, as the operators $\ICc^{\sf GZ}_{\cal P}$ and $\ICc^{\sf MR}_{\cal P}$ are generalizations of the operator-based characterisations by \citet{vanbesien2022analyzing} of the corresponding semantics.

\section{Conclusion}\label{sec:conclusion}
The main contributions of this paper are the definition of several ndaos for choice programs, the definition of the constructive stable operator, the characterisation of several existing semantics for various dialects of logic and choice programs and the introduction and study of postulates for choice programs. We provide a principled view of choice programs versus disjunctive programs.

This paper is subject to one restriction: we assume $\ICc^c_\mathcal{P}(x,y)\neq\emptyset$ for every interpretation $(x,y)$. In future work, we will generalize our results beyond this assumption. We also plan to study the complexity of the resulting semantics, device implementations and study other AFT-based semantics, such as the Kripke-Kleene and well-founded states and semi-equilibrium semantics \citep{DBLP:journals/corr/abs-2305-10846,heyninck2022non} and study other operators, e.g.\ inspired by \cite{pelov2007well}.

\nocite{cousot1979constructive}

\paragraph*{Acknowledgements}
I thank Bart Bogaerts and Wolfgang Faber for interesting discussions on this topic, and the reviewers for their dilligent reviews.
\bibliographystyle{named}
\bibliography{nondetAFT}

\end{document}